\pdfoutput=1
\documentclass[11pt]{article}

\usepackage{acl}

\usepackage{times}
\usepackage{latexsym}

\usepackage[T1]{fontenc}
\usepackage[utf8]{inputenc}

\usepackage{microtype}

\usepackage{times}
\usepackage{latexsym}
\usepackage{url}
\usepackage{enumitem}
\usepackage{booktabs}
\usepackage{multirow}
\usepackage{array}
\usepackage{graphicx}
\usepackage{subcaption}
\usepackage{amsmath}
\usepackage{amsfonts}
\usepackage{amsthm,amssymb,amsopn,bm}
\usepackage{color,soul}
\usepackage{verbatim}
\usepackage{tabulary}
\usepackage{footnote}
\usepackage{tabularx,booktabs}
\newcolumntype{Y}{>{\centering\arraybackslash}X}
\newcolumntype{P}[1]{>{\centering\arraybackslash}p{#1}}
\usepackage{xspace}
\usepackage{stmaryrd}
\usepackage{makecell}
\usepackage{stfloats}
\usepackage{booktabs}% http://ctan.org/pkg/booktabs
\newcommand{\tabitem}{~~\llap{\textbullet}~~}

\usepackage[T1]{fontenc}
\usepackage[utf8]{inputenc}

\usepackage{microtype}
\makesavenoteenv{tabular}
\usepackage{tabularx,booktabs}

\newcommand\bolden[1]{{\boldmath\bfseries#1}}

\newif\ifshorten % to make it 8 pages
\shortenfalse

\providecommand{\commentout}[1]{}

\title{Noisy Channel Language Model Prompting \\
for Few-Shot Text Classification}

\newcommand{\affilsup}[1]{\rlap{\textsuperscript{\normalfont#1}}}

\author{
    Sewon Min\affilsup{1,2},
    ~~~Mike Lewis,\affilsup{2}
    ~~Hannaneh Hajishirzi\affilsup{1,3},
    ~~Luke Zettlemoyer\affilsup{1,2} \\
    $^1$University of Washington \quad
    $^2$Facebook AI Research \quad
    $^3$Allen Institute for AI \\
    \texttt{\{sewon,hannaneh,lsz\}@cs.washington.edu} \qquad
    \texttt{mikelewis@fb.com} \\
}

\begin{document}
\maketitle
\begin{abstract}
    We introduce a noisy channel approach for language model prompting in few-shot text classification.
    Instead of computing the likelihood of the label given the input (referred as direct models), channel models compute the conditional probability of the input given the label, and are thereby required to explain every word in the input.
    We use channel models for recently proposed few-shot learning methods with no or very limited updates to the language model parameters, via either in-context demonstration or prompt tuning.
    Our experiments show that, for both methods, channel models significantly outperform their direct counterparts, which we attribute to their stability, i.e., lower variance and higher worst-case accuracy.
    We also present extensive ablations that provide recommendations for when to use channel prompt tuning instead of other competitive methods (e.g., direct head tuning): channel prompt tuning is preferred when the number of training examples is small, labels in the training data are imbalanced, or generalization to unseen labels is required.
    %We introduce a generative approach for language model prompting for few-shot text classification. Instead of computing the likelihood of the label given the input, we compute the conditional probability of the input given the label, using a causal language model without updating its parameters. Our experiments show that, with recently proposed few-shot learning methods with no or lightweight updates of the language model parameters---in-context demonstration and prompt tuning, generative models significantly outperform their discriminative counterparts. Our extensive ablations provide recommendations between generative prompt tuning and other competitive baselines (e.g., discriminative head tuning): generative prompt tuning is preferred when the number of training examples is small, labels in the training data are imbalanced, and generalization to unseen labels is required.
\end{abstract}

\section{Introduction}\label{sec:intro}%Few-shot learning requires a model to learn a task given a handful of training examples. Recently, p
Prompting large language models, by prepending natural language text or continuous vectors (called {\em prompts}) to the input, has shown to be promising in few-shot learning~\citep{brown2020language}. Prior work has proposed methods for finding better prompt
%better methods for constructing a prompt
~\citep{shin2020autoprompt,li2021prefix,lester2021power} or better scoring of the output from the model~\citep{zhao2021calibrate,holtzman2021surface}.
%These studies {\em directly} predict tokens: the (perhaps renormalized or finetuned) language model scores particular target tokens to determine the end task predictions.  
These studies {\em directly} predict target tokens to determine the prediction for an end task.
Despite promising results, they can be unstable with high variance across different verbalizers (text expression for labels) and seeds,
%makes them unlikely to be useful in practical few-shot learning
and the worst-case performance is often close to random~\citep{perez2021true,lu2021fantastically}.

\begin{figure}[t]
\resizebox{1.1\columnwidth}{!}{\includegraphics[trim={17.2cm 1cm 10.5cm 0},clip]{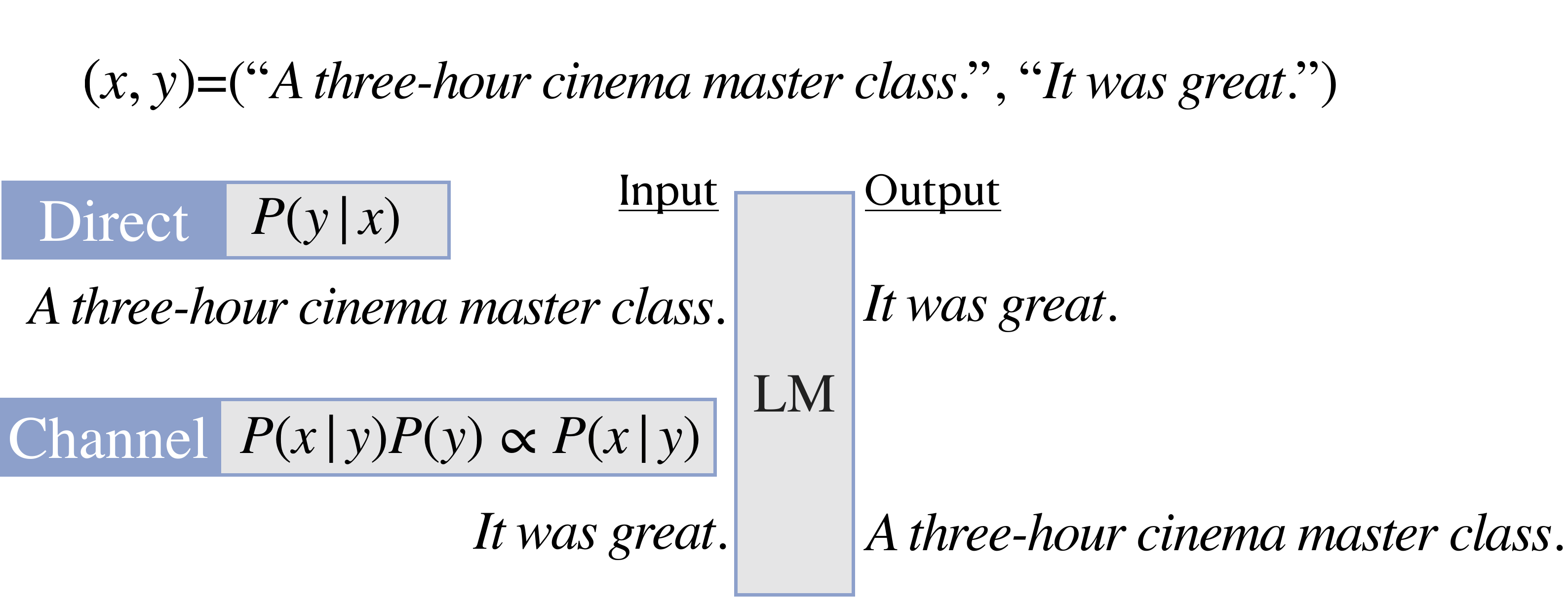}}\vspace{-.1cm}
\caption{An illustration of the direct model and the channel model for language model prompting in the sentiment analysis task.}\label{fig:teaser}
\end{figure}

In this paper, we introduce alternative {\em channel} models for prompted few-shot text classification with large language models, inspired by noisy channel models in machine translation~\citep{brown1993mathematics,koehn2003statistical,yu2016neural,yee2019simple} and their extensions to other tasks~\citep{yogatama2017generative,lewis2018generative}.
Unlike {\em direct} models that compute the conditional probability of the label token given the input, channel models compute the conditional probability of the input given the output (Figure~\ref{fig:teaser}).
Intuitively, channel models are required to explain every word in the input, potentially amplifying training signals in the low data regime.
We study the impact of channel models for language model prompting where the parameters of the language model are frozen.
In particular, 
we compare channel models with their direct counterparts for (1) demonstration methods, either concatenation-based~\citep{brown2020language} or our proposed, ensemble-based (Section~\ref{subsubsec:ensemble}), and (2) prompt tuning~\citep{lester2021power}.

Our experiments on eleven text classification datasets show that channel models outperform their direct counterparts by a large margin.
We attribute the strong performance of channel models to their stability: they have lower variance and significantly higher worst-case accuracy then their direct counterparts over different verbalizers and seeds.
We additionally find a direct model with {\em head tuning}---tuning the LM head while freezing other parameters---is surprisingly effective, often outperforming direct models with other forms of tuning.
While different methods are preferred given different conditions, the channel model with prompt tuning (denoted as channel prompt tuning) significantly outperforms all direct baselines when (1) the training data is imbalanced, or (2) generalization to unseen labels is required.

In summary, our contributions are three-fold:
\vspace{-.1cm}
\begin{enumerate}\setlength\itemsep{-.08cm}
    \item We introduce a noisy channel approach for language model prompting in few-shot text classification, showing that they significantly outperform their direct counterparts for both demonstration methods and prompt tuning.
    \item We find particularly strong performance of channel models over direct models when the training data is imbalanced or generalization to unseen labels is required.
    \item Based on extensive ablations, we provide recommendations between different models (direct vs. channel and prompt tuning vs. head tuning) based on given conditions such as the target task, the size of training data, the number of classes, the balance between labels in the training data, and whether generalization to unseen labels is required.
\end{enumerate}

\section{Related Work}\label{sec:related}\subsection{Channel Model}
Let $x$ and $y$ be the input and the output, respectively. The most widely used models, denoted as {\em direct} models, compute $P(y|x)$.
In contrast, {\em noisy channel} models maximize $P(x|y)P(y)$~\citep{shannon1948mathematical,brown1993mathematics}.\footnote{
    We follow \citet{yu2016neural,yee2019simple} in using the terms direct models and channel models. They are often referred as discriminative models and generative models in prior work~\citep{yogatama2017generative,lewis2018generative}.
    In principle, these two distinctions are not always equivalent, e.g., a model that computes $P(x,y)=P(y|x)P(x)$ is generative but not a channel model.
}
While the noisy channel approach has been the most successful in machine translation~\citep{yamada2001syntax,koehn2003statistical,yu2016neural,yee2019simple}, it has also been studied in more general NLP tasks. Prior work provides a theoretical analysis that channel models approach their asymptotic errors more rapidly than their direct counterparts~\citep{ng2002discriminative}, and empirically shows that channel models are more robust to distribution shift in text classification~\citep{yogatama2017generative} or question answering~\citep{lewis2018generative}, and in a few-shot setup~\citep{ding2019latent}.

In this paper, we explore channel models using a large language model on a wide range of text classification tasks, focusing on prompt-based few-shot learning.

\subsection{Few-shot Learning}

%Few-shot learning, where a model learns a task given limited number of labeled examples, is an important area of research both in an academic and a practical setting.
Prior work in few-shot learning has used different approaches, including semi-supervised learning with data augmentation or consistency training~\citep{miyato2017adversarial,clark2018semi,xie2020unsupervised,chen2020mixtext} and meta learning~\citep{finn2017model,huang2018natural,bansal2020self}.
Recent work has introduced {\em prompting} (or {\em priming}) of a large language model. For example, \citet{brown2020language} proposes to use a concatenation of training examples as a demonstration, so that when it is prepended to the input and is fed to the model, the model returns the output following the pattern in the training examples.
This is especially attractive as it eliminates the need for updating parameters of the language model, which is often expensive and impractical.
Subsequent work proposes alternative ways of scoring labels through better model calibration~\citep{zhao2021calibrate,holtzman2021surface}, or learning better prompts, either in a discrete space~\citep{shin2020autoprompt,jiang2020can,gao2021making} or in a continuous space~\citep{li2021prefix,lester2021power,liu2021gpt,zhong2021factual,qin2021learning}.
Almost all of them are direct models, computing the likelihood of $y$ given $x$ with the prompts.

Our work is closely related to two recent papers.
\citet{tam2021improving} studies a label-conditioning objective for masked language models; although this is not strictly a generative channel model, conditioning on the output $y$ is similar to our work.
However, they are still optimizing a discriminative objective, and inference at test time is the same as with the direct model.
%Moreover, it works with finetuning the entire language model, which is not possible for our zero- and few-shot evaluations.
\citet{holtzman2021surface} explores
zero-shot models that compute the probability of $x$ given $y$ based on Pointwise Mutual Information, but with a restriction that the input and the output are interchangeable.
To the best of our knowledge, our work is the first that uses a noisy channel model for few-shot language model prompting for classification, and also the first to draw the connection with the noisy channel literature.

\section{Formulation}\label{sec:approach}
\newcommand{\discr}{$P_\mathrm{LM}(v(c_i)|x)$}
\newcommand{\discrplus}{$\frac{P_\mathrm{LM}(v(c_i)|x)}{P_\mathrm{LM}(v(c_i)|\texttt{NULL})}$}
\newcommand{\gen}{$P_\mathrm{LM}(x|v(c_i))$}

\newcommand{\discrConcat}{$P_\mathrm{LM}(v(c_i)|x^1, v(c^1)...x^k, v(c^k), x)$}
\newcommand{\discrplusConcat}{$\frac{P_\mathrm{LM}(v(c_i)|x^1, v(c^1)...x^k, v(c^k), x)}{P_\mathrm{LM}(v(c_i)|x^1, v(c^1)...x^k, v(c^k), \texttt{NULL})}$}
\newcommand{\genConcat}{$P_\mathrm{LM}(x|x^1, v(c^1)...x^k, v(c^k), v(c_i))$}
\newcommand{\discrEns}{$\Pi_{j=1}^K P_\mathrm{LM}(v(c_i)|x^j, v(c^j), x)$}
\newcommand{\discrplusEns}{$\Pi_{j=1}^K \frac{P_\mathrm{LM}(v(c_i)|x^j, v(c^j), x)}{P_\mathrm{LM}(v(c_i)|x^j, v(c^j), \texttt{NULL})}$}
\newcommand{\genEns}{$\Pi_{j=1}^K P_\mathrm{LM}(x|v(c^j), x^j, v(c_i))$}

\begin{table*}[t]
    \centering \small
    \begin{tabular}{
            l @{\hspace{2em}} l @{\hspace{2em}} l @{\hspace{2em}} l}
        \toprule
            Method & Zero-shot & Concat-based Demonstrations & Ensemble-based Demonstrations \\
        \midrule
        \vspace{.3cm}
            \makecell[l]{Direct %\\  $P(c_i|x)$
            } & \discr & \discrConcat & \discrEns \\
        %\midrule
        \vspace{.3cm}
            \makecell[l]{Direct++ %\\ $\frac{P(c_i|x)}{P(c_i|\texttt{NULL})}$
            } & \discrplus & \discrplusConcat & \discrplusEns\\
        %\midrule
            \makecell[l]{Channel %\\  $P(x|c_i)$
            } & \gen & \genConcat & \genEns \\
        \bottomrule
    \end{tabular}
    \caption{
        Comparison of zero-shot, concat-based demonstrations, and ensemble-based demonstrations (Section~\ref{subsec:no-fintune-setup}).
        $\{(x^j,c^j)\}_{j=1}^K$ is training data and $v$ is the verbalizer. %, and $C(\{a^j, b^j\}_{j=1}^K)$ is a concatenation of $a^1, b^1, \cdots, a^K, b^K$.
    }\label{tab:method}
\end{table*}

We focus on text classification tasks. The goal is to learn a task function $f: \mathcal{X} \xrightarrow{} \mathcal{C}$, where $\mathcal{X}$ is the set of all natural language texts and $\mathcal{C}=\{c_1...c_m\}$ is a set of labels. We consider three formulations.

\vspace{.12cm}
\noindent
\textbf{Direct} computes distributions of labels $c_i \in \mathcal{C}$ given the input $x \in \mathcal{X}$: $P(c_i|x)$. This is the most widely used method in modern neural networks.

\vspace{.12cm}
\noindent
\textbf{Direct++} is a stronger direct model that computes $\frac{P(c_i|x)}{P(c_i|\texttt{NULL})}$ instead of $P(c_i|x)$, following the method from \citet{holtzman2021surface} and the non-parametric method from \citet{zhao2021calibrate}.
This approach is motivated by the fact that language models can be poorly calibrated and suffer from competition between different strings with the same meaning. This approach is used for the demonstration methods in Section~\ref{subsec:no-fintune-setup}.

\vspace{.12cm}
\noindent
\textbf{Channel} %computes the joint probability of $(x, c_i)$ which we reparameterize to $P(x|c_i)P(c_i)$. We consider classification tasks with an assumption of uniform distributions over labels, thus $P(c_i)=\frac{1}{|\mathcal{C}|}$ and we mainly care about $P(x|c_i)$.
uses Bayes' rule to reparameterize $P(c_i|x)$ as $\frac{P(x|c_i)P(c_i)}{P(x)}$.
As we are generally interested in $\mathrm{argmax}_{c_i \in \mathcal{C}}\frac{P(x|c_i)P(c_i)}{P(x)}$ and $P(x)$ is independent from $c_i$, it is sufficient to model $P(x|c_i)P(c_i)$.
We assume $P(c_i)=\frac{1}{|\mathcal{C}|}$ and only compute $P(x|c_i)$.

\section{Method}\label{sec:method}We explore direct and channel models using a causal language model (LM) $P_\mathrm{LM}$ that gives the conditional probability of the text $y$ when followed by $x$.
More precisely, given the text $x=x_1...x_{t_x}$ and $y=y_1...y_{t_y}$ ($x_1...x_{t_x},y_1...y_{t_y} \in \mathcal{V}$, where $\mathcal{V}$ is the vocabulary set),
$P_\mathrm{LM}(y|x)$ indicates
$\Pi_{t'=1}^{t_y} P_\mathrm{LM}(y_{t'}|x_1...x_{t_x}y_1...y_{t'-1})$.\footnote{
    In practice, we use length normalization that was found to be effective by \citet{holtzman2021surface}.
    %over the log likelihood of $y_1...y_{t_x}$---that was found to be effective by \citet{holtzman2021surface}. 
    %This matters only for a discriminative model, only when the lengths of the verbalized text vary across classes.
}

When learning a task function $f: \mathcal{X} \xrightarrow{} \mathcal{C}$, we also assume a pre-defined {\em verbalizer} $v: \mathcal{C} \xrightarrow{} \mathcal{X}$ which maps each label into a natural language expression.
\newcommand{\x}{``A three-hour cinema master class''}
\newcommand{\p}{``It was great''}
\newcommand{\n}{``It was terrible''}
For example, if the task is sentiment analysis with $\mathcal{C}=\{c^+, c^-\}$, an example input text $x$ would be \x\ and an example $v$ would have $v(c^+)=$\p\ and $v(c^-)=$\n.
%We also assume $\mathcal{D}$ that consists of two training examples, (``Too silly to take seriously'', $c^-$) and (``The film is one of the year's best'', $c^+$).
%
In a few-shot setup, we are also given a set of $K$ training examples $\mathcal{D}=\{(x^1, c^1), \cdots, (x^K, c^K)\}$.

We are interested in methods where there are no trainable parameters (Section~\ref{subsec:no-fintune-setup}) or the number of trainable parameters is very small, typically less than 0.01\% of the total (Section~\ref{subsec:finetune-setup}). This follows prior observations that updating and saving a large number of parameters for every task is expensive and often infeasible~\citep{rebuffi2017learning,houlsby2019parameter,lester2021power}.

\subsection{Demonstration methods}\label{subsec:no-fintune-setup}
In demonstration methods, there are no trainable parameters. We explore three ways of making a prediction, as summarized in Table~\ref{tab:method}. %, two of which are from \citet{brown2020language} and the third from this paper. %See Table~\ref{tab:method} for an overview.

\subsubsection{Zero-shot}\label{subsubsec:zeroshot-setup}
We follow \citet{brown2020language} in computing $P(c_i|x)$ and $P(x|c_i)$ as $P_\mathrm{LM}(v(c_i)|x)$ and $P_\mathrm{LM}(x|v(c_i))$, respectively.
For example, given $x=$\x, the direct model compares the probabilities of \p\ and \n\ when following \x, while the channel model considers the probabilities of \x\ when following \p\ or \n.

\subsubsection{Concat-based demonstrations}
We follow the few-shot learning method in \citet{brown2020language}.
The key idea is to prepend a concatenation of $K$ training examples to the input so that a language model can learn the task setup from the input.
The original method was used for a direct model, but can be naturally extended for a channel model.
Concretely, $P(c_i|x)$ in direct models is obtained via $P_\mathrm{LM}(v(c_i)|x^1, v(c^1), \cdots, x^K, v(c^K), x)$, and $P(x|c_i)$ in channel models is obtained via $P_\mathrm{LM}(x|v(c^1), x^1, \cdots, v(c^K), x^K, v(c_i))$.

\subsubsection{Ensemble-based demonstrations}\label{subsubsec:ensemble}

We propose a new method as an alternative to the concat-based method, which we find to be a stronger direct model. 
Instead of concatenating $K$ training examples as one sequence and getting output probabilities from an LM once, we obtain output probabilities from an LM $K$ times conditioned on one training example at a time, and multiply the resulting probabilities.
Specifically, $P(c_i|x)$ is computed via $\Pi_{j=1}^K P_\mathrm{LM}(v(c_i)|x^j, v(c^j), x)$ and $P(x|c_i)$ is computed via
$\Pi_{j=1}^K
P_\mathrm{LM}(x|v(c^j), x^j, v(c_i))$.
This method also reduces the memory consumption---the concat-based method uses $O(K^2)$ while this method uses $O(K)$---and eliminates the dependency on the ordering of training examples, which has been shown to significantly impact the model performance~\citep{zhao2021calibrate,lu2021fantastically}.

\begin{figure*}[t]
\resizebox{1.02\textwidth}{!}{\includegraphics{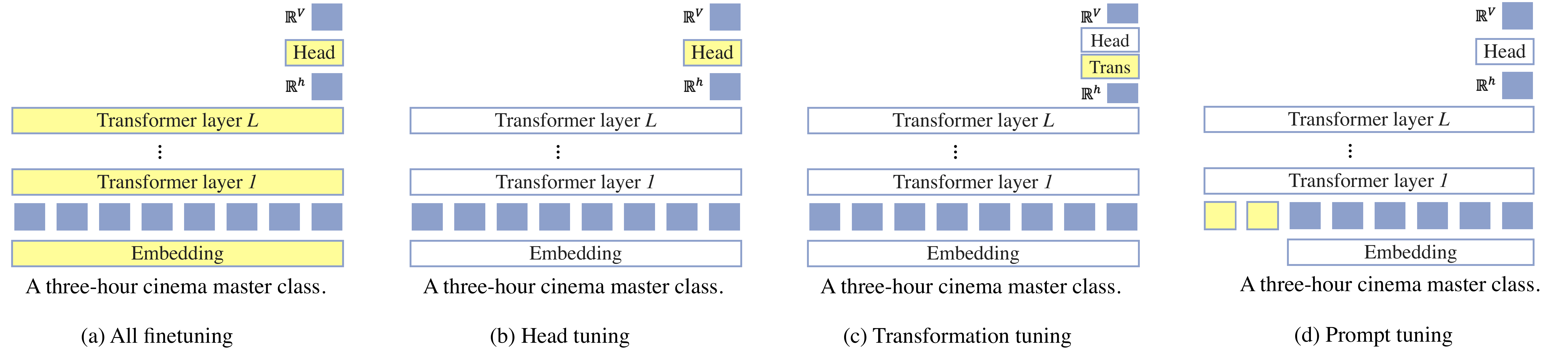}}
\caption{Different finetuning methods, which compute the distributions of the next token given \x. Yellow and white boxes are trainable and frozen parameters, respectively.
$h$ and $V$ denote the hidden dimension of the LM and the vocabulary size of $v(c_1)...v(c_m)$, respectively.
{\em All finetuning} is a typical finetuning method that updates all parameters of the LM (illustrated as a reference). {\em Head tuning}, {\em Transformation tuning} and {\em Prompt tuning} 
are described in Section~\ref{subsec:finetune-setup}; they update a very limited number of parameters.}
\label{fig:finetune}
\end{figure*}

\subsection{Tuning methods}\label{subsec:finetune-setup}
We also explore methods that tune a very limited number of model parameters, as summarized in Figure~\ref{fig:finetune}. We study head tuning (Section~\ref{subsec:head-tuning}) and transformation tuning (Section~\ref{subsec:transform-tuning}) for direct models. We also consider prompt tuning (Section~\ref{subsec:prompt-tuning}) for both direct and channel models, which we refer as direct prompt tuning and channel prompt tuning, respectively.
All models share the same input-output interface with the zero-shot setup in Table~\ref{tab:method} during training and inference.

\subsubsection{Head tuning}~\label{subsec:head-tuning}
Head tuning finetunes the head---the matrix in the LM which transforms the hidden representation from the last transformer layer to the logit values. Let $\mathbf{O} \in \mathbb{R}^{|\mathcal{V}| \times h}$ be the head and $\mathbf{h}_x \in \mathbb{R}^h$ be the hidden representations from the last transformer layer given $x$,
$P_\mathrm{LM}(v_i|x)$ for a token $v_i \in \mathcal{V}$ is computed via an $i$-th element of $\mathrm{Softmax}(\mathbf{O} \mathbf{h}_x)$.
We finetune $\mathbf{O}$ while freezing all other parameters of the LM.
%\ifshorten{
Although $\mathbf{O}$ is tied with the embedding matrix of the LM during language model pretraining, we separate them during head tuning.\footnote{This is different from head tuning from prior work, e.g., \citet{le-scao-rush-2021-many}, which finetunes $\tilde{P}_\mathrm{LM}$ and uses a separate, randomly initialized head instead of the LM head.}
%}\else{Although $\mathbf{O}$ is tied with the embedding matrix of the LM during language model pretraining, we separate them during finetuning. The number of trainable values is $|v(\mathcal{C})|h$ where $v(\mathcal{C})$ denotes vocabularies in $v(c_1)...v(c_m)$.\footnote{This is different from head tuning from prior work, e.g., \citet{le-scao-rush-2021-many}, which finetunes $\tilde{P}_\mathrm{LM}$ and uses a seperate, randomly initialized head instead of the LM head.}
%}\fi

\subsubsection{Transformation tuning}~\label{subsec:transform-tuning}
As an alternative to head tuning, we transform $\mathbf{O}$ with a new transformation matrix $\mathbf{U} \in \mathbb{R}^{h \times h}$.
Specifically, $P_\mathrm{LM}(v_i|x)$ for a token $v_i \in \mathcal{V}$ is computed via an $i$-th element of $\mathrm{Softmax}(\mathbf{O}\mathbf{U}\mathbf{h}_x)$.
We train $\mathbf{U}$, initialized from an identity matrix, and freeze other parameters including $\mathbf{O}$.
%The number of trainable values is $h^2$.

\subsubsection{Prompt tuning}~\label{subsec:prompt-tuning}
Prompt tuning is the method that has recently gathered much attention~\citep{li2021prefix,lester2021power,liu2021gpt}.
The key idea is to consider the LM as a black-box model and instead learn continuous prompt embeddings.
We follow the method from \citet{lester2021power} where $n$ prompt tokens $u_1...u_n$ are prepended to the input, and the embeddings of $u_1...u_n$ are learned.
In other words, direct models compute $P(c_i|x)=P_\mathrm{LM}(v(c_i)|u_1...u_n, x)$, and channel models compute $P(x|c_i)=P_\mathrm{LM}(x|u_1...u_n, v(c_i))$.
The parameters in the LM are frozen except the embeddings of $u_1...u_n$.\footnote{
    This is different from prompt tuning in~\citet{gao2021making,liu2021gpt}
    which jointly trains prompt embeddings and the parameters of the LM.}
%so that the number of trainable values is $nh$.
\section{Experimental Setup}\label{sec:exp-setup}\begin{table}[t]
    \centering \small
    \begin{tabular}{l l r }
        \toprule
           Dataset & Task & $|\mathcal{C}|$\\
        \midrule
            SST-2 & Sentiment analysis (movie) & 2 \\
            SST-5 & Sentiment analysis (movie) & 5 \\
            MR & Sentiment analysis (movie) & 2 \\
            CR & Sentiment analysis (electronics) & 2 \\
            Amazon & Sentiment analysis (Amazon) & 5 \\
            Yelp & Sentiment analysis (Yelp) & 5 \\
            TREC & Question classification (answer type) & 6 \\
            AGNews & News classification (topic) & 4 \\
            Yahoo & Question classification (topic) & 10 \\
            DBPedia & Ontology classification & 14 \\
            Subj & Subjectivity classification & 2 \\
        \bottomrule
    \end{tabular}
    \caption{Datasets used for experiments. $|\mathcal{C}|$ denotes the number of classes.See Appendix~\ref{app:verb} for samples.}\label{tab:dataset}
\end{table}

%\begin{table}[t]
%    \centering \small
%    \begin{tabular}{l l r }
%        \toprule
%           Dataset & Task & $|\mathcal{C}|$\\
%        \midrule
%            SST-2 & Sentiment analysis (movie) & 2 \\
%            SST-5 & Sentiment analysis (movie) & 5 \\
%            MR & Sentiment analysis (movie) & 2 \\
%            CR & Sentiment analysis (electronics) & 2 \\
%            Amazon & Sentiment analysis (Amazon) & 5 \\
%            Yelp & Sentiment analysis (Yelp) & 5 \\
%            TREC & Question classification (answer type) & 6 \\
%            AGNews & News classification (topic) & 4 \\
%            Yahoo & Question classification (topic) & 10 \\
%            DBPedia & Ontology classification & 14 \\
%            Subj & Subjectivity classification & 2 \\
%        \bottomrule
%    \end{tabular}
%    \caption{Datasets used for experiments. $|\mathcal{C}|$ denotes the number of classes.See Appendix~\ref{app:verb} for samples.}\label{tab:dataset}
%\end{table}

\subsection{Datasets}\label{subsec:data}

We report results for eleven text classification datasets, following \citet{zhang2015character} and \citet{gao2021making}: SST-2~\citep{socher2013recursive}, SST-5~\citep{socher2013recursive}, MR~\citep{pang2005seeing}, CR~\citep{hu2004mining}, Amazon~\citep{mcauley2013hidden}, Yelp~\citep{zhang2015character}, TREC~\citep{voorhees2000building}, AGNews~\citep{zhang2015character}, Yahoo~\citep{zhang2015character}, DBPedia~\citep{lehmann2015dbpedia} and Subj~\citep{pang2004sentimental}.
The datasets include a varied number of classes per task, from 2 to 14. % (Table~\ref{tab:dataset}).
See Table~\ref{tab:data-samples} in Appendix~\ref{app:verb} for dataset samples.

\commentout{
    \begin{table}[t]
        \centering \small
        \begin{tabular}{l l l @{\hspace{1em}} r}
            \toprule
                Model & \# of parameters & \# of layers & $h$ \\
            \midrule
                Small & 117M & 12 & 768 \\
                Medium & 345M & 24 & 1024 \\
                Large & 762M & 36 & 1280 \\
                X-Large & 1542M & 48 & 1600 \\
            \bottomrule
        \end{tabular}
        \caption{
            Sizes of different GPT-2 models. $h$ denotes a hidden dimension of the model.
            We use a {\em Large} model if not specified otherwise.
        }\label{tab:lm}
    \end{table}
}

\newcommand{\nofinetune}{
    SST-2 & 63.0/51.1 & 80.3/76.9 & 77.1/74.8 &
    58.9/50.6 & 66.8/51.7 & \textbf{85.0}/\textbf{83.1} & 
    57.5/50.9 & 79.7/68.0 & 77.5/59.5
    \\
    SST-5 & 27.5/24.4 & 33.3/28.8 & 29.2/27.7 & 
    27.6/23.0 & 23.7/14.4 & \textbf{36.2}/\textbf{32.7} &
    25.6/23.2 & 33.8/23.3 & 33.6/30.2
    \\
    MR & 61.7/50.3 & 77.4/73.2 & 74.3/69.3 & 
    56.4/50.0 & 60.2/50.5 & \textbf{80.5}/\textbf{76.8} &
    58.8/50.0 & 76.8/60.1 & 76.1/60.0
    \\
    CR & 59.2/50.0 & 77.9/69.7 & 65.8/60.2 & 
    54.7/50.0 & 66.8/50.0 & \textbf{80.8}/\textbf{74.8} & 51.0/50.0 & 72.8/54.6 & 79.7/69.3
    \\
    Amazon & 31.2/22.4 & 37.6/35.0 & 37.1/31.6 & 
    33.0/21.4 & \textbf{40.8}/35.7 & 39.4/34.3 &
    31.7/23.1 & 39.8/32.0 & 40.4/\textbf{36.2}
    \\
    Yelp & 33.2/25.6 & 36.8/31.8 & 38.0/31.9 & 
    32.6/23.3 & 38.5/31.6 & 39.8/36.5 &
    31.4/23.6 & 39.2/29.6 & \textbf{41.5}/\textbf{38.5}
    \\
    AGNews & 59.8/47.8 & 59.9/44.0 & 61.8/59.7 & 
    34.0/25.0 & 51.2/34.4 & 68.5/60.6 &
    51.9/34.2 & 73.1/58.6 & \textbf{74.3}/\textbf{69.3}
    \\
    TREC & 38.7/26.0 & 27.7/12.6 & 30.5/19.4 &
    27.2/9.4 & 31.6/13.0 & \textbf{42.0}/\textbf{26.8} &
    32.1/13.0 & 22.9/9.8 & 31.5/23.8
    \\
    Yahoo & 20.7/17.8 & 35.3/28.7 & 48.7/48.1 &
    13.0/10.0 & 29.6/19.4 & 56.2/52.3 &
    16.6/10.7 & 50.6/46.5 & \textbf{58.6}/\textbf{57.4}
    \\
    DBPedia & 32.3/18.6 & 37.6/30.4 & 51.4/42.7 & 
    32.5/7.1 & 71.1/55.2 & 58.5/40.0 &
    46.8/17.1 & \textbf{72.6}/55.7 & 64.8/\textbf{57.0}
    \\
    Subj & 51.0/49.9 & 52.0/48.8 & 57.8/\textbf{51.5} & 
    53.7/49.9 & 56.9/50.0 & \textbf{60.5}/40.8 &
    51.6/49.6 & 52.2/41.8 & 52.4/46.9
    \\
    \midrule
    Avg. & 43.5/34.9 & 50.5/43.6 & 52.0/47.0 &
    38.5/29.1 & 48.8/36.9 & \textbf{58.9}/\textbf{50.8} &
    41.4/31.4 & 55.8/43.6 & 57.3/49.8
    \\
}

\newcommand{\nofinetuneold}{
    SST-2 & 63.0/51.1 & 80.3/76.9 & 77.1/74.8 &
    58.6/50.6 & 64.4/51.1 & \textbf{85.1}/\textbf{81.9} & 
    57.5/50.9 & 79.7/68.0 & 77.5/59.5
    \\
    SST-5 & 27.5/24.4 & 33.3/28.8 & 29.2/27.7 & 
    29.0/18.5 & 24.2/14.4 & \textbf{35.1}/\textbf{29.3} &
    25.6/23.2 & 33.8/23.3 & 33.6/30.2
    \\
    MR & 61.7/50.3 & 77.4/\textbf{73.2} & 74.3/69.3 & 
    56.7/50.0 & 60.7/50.1 & \textbf{80.4}/73.1 &
    58.8/50.0 & 76.8/60.1 & 76.1/60.0
    \\
    CR & 59.2/50.0 & 77.9/69.7 & 65.8/60.2 & 
    55.2/50.0 & 67.4/50.0 & \textbf{81.2}/\textbf{74.5} & 51.0/50.0 & 72.8/54.6 & 79.7/69.3
    \\
    Amazon & 31.2/22.4 & 37.6/35.0 & 37.1/31.6 & 
    \\
    Yelp & 33.2/25.6 & 36.8/31.8 & 38.0/31.9 & 
    \\
    TREC & 38.7/\textbf{26.0} & 27.7/12.6 & 30.5/19.4 &
    27.8/9.4 & 32.6/13.0 & \textbf{40.3}/20.2 &
    32.1/13.0 & 22.9/9.8 & 31.5/23.8
    \\
    AGNews & 59.8/47.8 & 59.9/44.0 & 61.8/59.7 & 
    34.8/25.0 & 50.5/27.2 & 68.1/57.6 & 51.9/34.2 & 73.1/58.6 & \textbf{74.3}/\textbf{69.3}
    \\
    Yahoo & 20.7/17.8 & 35.3/28.7 & 48.7/48.1 &
    \\
    DBPedia & 32.3/18.6 & 37.6/30.4 & 51.4/42.7 & 
    \\
    Subj & 51.0/49.9 & 52.0/48.8 & 57.8/51.5 & 
    34.8/25.0 & 50.5/27.2 & 68.1/57.6 & 51.9/34.2 & 73.1/58.6 & \textbf{74.3}/\textbf{69.3}
    \\
    \midrule
    Avg. & 43.5/34.9 & 50.5/43.6 & 52.0/47.0
    \\
}

\newcommand{\zeroDis}{
    63.0/51.1 & % 63.0$_{10.5}$/51.1 &
    27.5/24.4 & % 27.5$_{2.2}$/24.4 &
    61.7/50.3 & % 61.7$_{9.6}$/50.3 &
    59.2/50.0 & % 59.2$_{7.8}$/50.0 &
    31.2/22.4 & % 31.2$_{6.7}$/22.4 &
    33.2/25.6 & % 33.2$_{7.3}$/25.6 & %
    38.7/26.0 & % 38.7$_{8.0}$/26.0 & %
    59.8/47.8 & % 59.8$_{11.2}$/47.8 & %
    20.7/17.8 & % 20.7$_{2.6}$/17.8 & %
    32.3/18.6 & % 32.3$_{9.1}$/18.6 & %
    51.0/49.9 & % 51.0$_{1.8}$/49.9 & %
    43.5/34.9
}
\newcommand{\zeroDisP}{
    80.3/76.9 & %80.3$_{2.3}$/76.9 &
    33.3/28.8 & % 33.3$_{3.7}$/28.8 &
    77.4/73.2 & % 77.4$_{2.5}$/73.2 &
    77.9/69.7 & % 77.9$_{7.0}$/69.7
    37.6/35.0 & % 37.6$_{2.1}$/35.0
    36.8/31.8 & % 36.8$_{3.9}$/31.8
    27.7/12.6 & % 27.7$_{11.4}$/12.6
    59.9/44.0 & % 59.9$_{31.2}$/44.0
    35.3/28.7 & % 35.3$_{4.8}$/28.7
    37.6/30.4 & % 37.6$_{4.2}$/30.4
    52.0/48.8 & % 52.0$_{3.4}$/48.8
    50.5/43.6
}
\newcommand{\zeroGen}{
    77.1/74.8 & % 77.1$_{2.4}$/74.8 &
    29.2/27.7 & % 29.2$_{1.5}$/27.7 &
    74.3/69.3 & % 74.3$_{3.1}$/69.3 &
    65.8/60.2 & % 65.8$_{5.7}$/60.2 &
    37.1/31.6 & % 37.1$_{3.2}$/31.6 &
    38.0/31.9 & % 38.0$_{3.5}$/31.9 & %
    30.5/19.4 & % 30.5$_{11.1}$/19.4 & %
    61.8/59.7 & % 61.8$_{1.5}$/59.7 & %
    48.7/48.1 & % 48.7$_{0.7}$/48.1 & %
    51.4/42.7 & %  51.4$_{6.1}$/42.7 & %
    57.8/51.5 & % 57.8$_{4.2}$/51.5 & %
    52.0/47.0
}

\begin{table*}[t]
    \centering  \footnotesize
    \setlength{\tabcolsep}{0.5em}
    \begin{tabular}{
            l @{\hspace{1.2em}}
            ccc @{\hspace{1.2em}}
            ccc @{\hspace{1.2em}}
            ccc
        }
        \toprule
            \multirow{2}{*}{Data} & 
            \multicolumn{3}{c}{Zero-shot {\em (4 runs)}} &
            \multicolumn{3}{c}{Concat-based {\em (20 runs)}} &
            \multicolumn{3}{c}{Ensemble-based {\em (20 runs)}} \\
        \cmidrule(lr){2-4} \cmidrule(lr){5-7} \cmidrule(lr){8-10}
            %& Dis & Dis++ & Gen & Dis & Dis++ & Gen & Dis & Dis++ & Gen \\
            & Direct & Direct++ & Channel
            & Direct & Direct++ & Channel
            & Direct & Direct++ & Channel
            \\
        \midrule    
            \nofinetune
        \bottomrule
    \end{tabular}
    \caption{
        Results from demonstration methods. All with GPT-2 Large.
        %{\em Dis}, {\em Dis++} and {\em Gen} denote a naive discriminative approach, a discriminative approach from \citet{holtzman2021surface,zhao2021calibrate}, and a generative approach, respectively.
        Two numbers respectively indicate the average and the worst-case accuracy over different verbalizers (zero-shot and few-shot) and data seeds (few-shot). `Avg.' in the last row indicate the macro-average across all datasets.
    }\label{tab:main}
\end{table*}

\subsection{Training Data}\label{subsec:train-data}

For few-shot learning, we primarily use training set size $K=16$, but explore $K=\{4,16,64,\mathrm{Full}\}$ in the ablations.
We sample the $K$ examples uniformly from the {\em true} distribution of the training data.
We relax the assumption from prior work of an equal number of training examples per label~\citep{gao2021making,logan2021cutting}, for more realistic and challenging evaluation. 

%Apart from experiments with $K=\mathrm{full}$,
We follow all the hyperameters and details from prior work (Appendix~\ref{app:hyperparams}) which eliminates the need for a held-out validation set.
The very limited data is better used for training rather than validation, and cross-validation is less helpful when the validation set is extremely small~\citep{perez2021true}.

\subsection{Language Models}
We use GPT-2~\citep{radford2019language} for the LM.
We primarily use GPT-2 Large but also experiment with varying sizes (Small, Medium, Large and X-Large) for the ablations in Appendix~\ref{app:ablations}.
While we only experiment with GPT-2, our experiments are easily extendable to other causal language models.

\subsection{Evaluation}
We use accuracy as a metric for all datasets.

We experiment with 4 different verbalizers (taken from \citet{gao2021making}; full list provided in Appendix~\ref{app:verb}), 5 different random seeds for sampling training data, and 4 different random seeds for training.
%This means we have (1) 4 runs for a zero-shot setup (as data seeds and train seeds do not matter), (2) 20 runs for demonstration methods (as train seeds do not matter), and (3) 80 runs for tuning methods.
We then report {\em Average accuracy} %, {\em Standard deviation}
and {\em Worst-case accuracy}.\footnote{We also report standard deviation and best-case accuracy in the Appendix.}
We consider the worst-case accuracy to be as important as the average accuracy given significantly high variance of few-shot learning models, as shown in previous work~\citep{zhao2021calibrate,perez2021true}.
The worst-case accuracy is likely of more interest in high-risk applications~\citep{Asri2016UsingML,guo2017calibration}.

Other implementation details are in Appendix~\ref{app:hyperparams}.
All experiments are reproducible from
\href{https://github.com/shmsw25/Channel-LM-Prompting}{\nolinkurl{github.com/shmsw25/Channel-LM-Prompting}}.

\section{Experimental Results}\label{sec:exp-result}This section reports results from demonstration methods (Section~\ref{subsec:main-no-finetune-result}), tuning methods (Section~\ref{subsec:main-finetune-result}) and ablations (Section~\ref{subsec:ablation-result}).
Discussion is provided in Section~\ref{sec:discuss}.

\subsection{Main Results: Demonstration Methods}\label{subsec:main-no-finetune-result}

Table~\ref{tab:main} shows the performance of demonstration methods.

\vspace{-.12cm}
\paragraph{Direct vs. Direct++}
Direct++ significantly outperforms the naive direct model across all setups, indicating that using $\frac{P(c_i|x)}{P(c_i|\texttt{NULL})}$ instead of $P(c_i|x)$ is highly beneficial as claimed by \citet{holtzman2021surface,zhao2021calibrate}.

\vspace{-.12cm}
\paragraph{Concat vs. Ensemble} Our proposed, ensemble-based method is better than the concat-based method in direct models, by 7\% absolute in the average accuracy and the worst-case accuracy, when macro-averaged across all datasets.

In contrast, the ensemble-based method is not always better in channel models; it is better only on the datasets with long inputs.
We conjecture that the ensemble-based method may suffer when labels in the training data are not balanced, which direct++ explicitly takes into account as described in~\citet{zhao2021calibrate}.

\vspace{-.12cm}
\paragraph{Direct++ vs. Channel} In a few-shot setting, channel models outperform direct models in almost all cases. The strongest channel model outperforms the strongest direct model by 3.1\% and 7.2\% absolute, in terms of the average accuracy and the worst-case accuracy, respectively.

Standard deviation and the best-case accuracy are reported in Table~\ref{tab:no-finetune-concat-full-result} and Table~\ref{tab:no-finetune-ensemble-full-result} in the Appendix.
They indicate strong performance of channel models can be attributed to their low variance. The highest best-case accuracy is achieved by direct++ on most datasets, but it has a higher variance, having lower average and the worst-case accuracy than channel models.

\vspace{-.12cm}
\paragraph{Zero-shot vs. Few-shot}
Performance of direct models sometimes degrades in a few-shot setting, which is also observed by prior work~\citep{zhao2021calibrate}. This is likely because demonstrations provided by the training data may cause the model to be miscalibrated and easily biased by the choice of demonstrations. However, channel models achieve few-shot performance that is significantly better than zero-shot methods on all datasets.

%\vspace{-.12cm}
%\paragraph{Comparisons to prior work}
%Our zero-shot numbers are comparable to those from \citet{zhao2021calibrate} and \citet{holtzman2021surface} on shared datasets. % (SST-2, SST-5, TREC, AGNews, DBPedia), However, there are cases where numbers from all three papers are significantly different.\footnote{For instance, for discriminative zero-shot, ours (Large), \citet{zhao2021calibrate} (X-Large) and \citet{holtzman2021surface} (Large) achieve 38.7, 24.0 and 21.4 on TREC, and 59.8, 44.0 and 69.8 on AGNews, respectively.} It is likely because of differences in pattern\footnote{A function that maps the input text $x$ to the input to be fed into the LM, following the notation from \citet{schick2020exploiting,le-scao-rush-2021-many}.} and verbalizers, and we did our best to eliminate such impact by not using a task-specific pattern and running multiple times with different verbalizers.

\subsection{Main Results: Tuning Methods}\label{subsec:main-finetune-result}

Table~\ref{tab:finetune} shows the performance of tuning methods.

\vspace{-.12cm}
\paragraph{Comparison when prompt tuning}
When using prompt tuning, channel models consistently outperform direct models by a large margin on all datasets. Improvements are 13.3\% and 23.5\% absolute in the average and the worst-case accuracy, respectively.

Standard deviation and the best-case accuracy are reported in Table~\ref{tab:finetune-full-result} in the Appendix.
Consistent with the findings in Section~\ref{subsec:main-no-finetune-result}, the strong performance of channel prompt tuning can be explained by the low variance of channel prompt tuning.
Direct prompt tuning often achieves higher best-case accuracy; however, due to its high variance, its overall accuracy is lower, with significantly lower worst-case accuracy.

\begin{table}[t]
    \centering \footnotesize
    \setlength{\tabcolsep}{0.4em}
    \begin{tabular}{
            l @{\hspace{1em}} cccc
        }
        \toprule
            %Data & Dis Head & Dis Trans & Dis Prompt & Gen Prompt \\
            \multirow{2}{*}{Data} & 
            \multicolumn{3}{c}{Direct} & Channel \\
            \cmidrule(lr){2-4} \cmidrule(lr){5-5}
            & Head & Trans & Prompt & Prompt \\
        \midrule    
            SST-2 &
            80.2/68.6 & 77.3/57.5 & 72.6/50.9 & \textbf{85.8}/\textbf{81.3} \\
            SST-5 &
            34.9/30.0 & 33.0/25.5 & 30.9/19.1 & \textbf{36.3}/\textbf{27.9} \\
            MR &
            73.7/\textbf{56.4} & 71.3/51.6 & 67.4/50.1 & \textbf{81.7}/78.0 \\
            CR &
            67.6/50.0 & 63.9/50.0 & 65.7/50.0 & \textbf{79.6}/\textbf{76.4} \\
            Amazon &
            34.5/28.8 & 32.1/18.2 & 31.2/20.0 & \textbf{43.4}/\textbf{39.2} \\
            Yelp &
            40.6/32.8 & 38.9/31.5 & 31.9/20.6 & \textbf{43.9}/\textbf{37.2} \\
            TREC &
            \textbf{54.1}/\textbf{42.4} & 48.0/31.0 & 35.9/13.0 & 37.1/20.8 \\
            AGNews &
            \textbf{74.1}/61.2 & 66.9/47.0 & 61.9/25.2 & 73.4/\textbf{63.9} \\
            Yahoo &
            39.1/31.4 & 33.8/23.0 & 27.4/15.7 & \textbf{54.0}/\textbf{46.7} \\ 
            DBPedia &
            49.3/37.5 & 42.4/28.6 & 41.8/9.9 & \textbf{67.7}/\textbf{52.9} \\
            Subj &
            \textbf{86.3}/\textbf{79.1} & 86.0/71.6 & 65.5/49.9 & 75.5/58.8 \\
        \midrule
            Avg. &
            57.7/47.1 & 54.0/39.6 & 48.4/29.5 & \textbf{61.7}/\textbf{53.0} \\
        \bottomrule
    \end{tabular}
    \caption{
        Performance of tuning methods with a limited number of trainable parameters.
        All methods use GPT-2 Large, and are run 80 times.
        %{\em Dis} and {\em Gen} indicate discriminative and generative, respectively.
        {\em Head}, {\em Trans}, {\em Prompt} indicate head tuning, transformation tuning and prompt tuning, respectively.
        We report the average / worst-case accuracies, separated by a slash.
        `Avg.' is the macro-average across all datasets.
    }\label{tab:finetune}
\end{table}

\vspace{-.12cm}
\paragraph{Head tuning vs. prompt tuning}
We find that head tuning is a very strong method, despite often being omitted as a baseline in prior work.
It significantly outperforms direct prompt tuning in all cases.
It also outperforms channel prompt tuning on some datasets, particularly significantly on TREC and Subj.
For these datasets, the task---finding the type of the answer to the question or identifying the subjectivity of the statement---is inherently different from language modeling, and likely benefits from directly updating the LM parameters, rather than using the LM as a black box.

Still, channel prompt tuning outperforms direct head tuning on most datasets. The largest gains are achieved on Yahoo and DBPedia. In fact, on these datasets, channel prompt tuning even outperforms {\em all finetuning}---finetuning all parameters of the LM---which achieves 48.9/43.8 on Yahoo and 66.3/50.4 on DBPedia. We conjecture that using $K=16$ on these datasets naturally requires generalization to unseen labels due to the large number of classes ($|\mathcal{C}|=10$ and $14$), where channel prompt tuning significantly outperforms direct models, as we show in Section~\ref{subsec:generalization-result}.

%\begin{figure*}[t]
%\resizebox{2.1\columnwidth}{!}{\includegraphics{figures/size_ablation.pdf}}
%\caption{\textbf{Varying the size of LMs} from GPT-2 Small to GPT-2 X-Large. The average accuracy (top) and the worst-case accuracy (bottom) are reported. All models are run 20 times (4 verbalizers and 5 data seeds). {\em Head} and {\em Prompt} indicate head tuning and prompt tuning, respectively. {Trends are consistent across different sizes of LM.}}
%\label{fig:size_ablation}
%\end{figure*}

\begin{figure*}[t]
\resizebox{2.1\columnwidth}{!}{\includegraphics{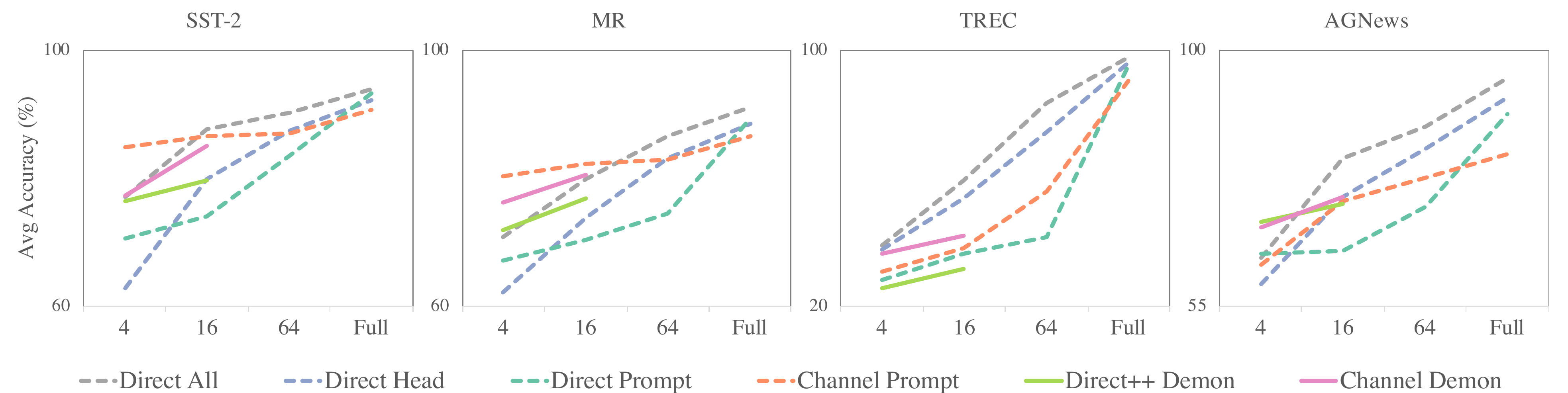}}
\caption{
    \textbf{Varying the number of training examples ($K$)}. All models use GPT-2 Large.
    %{\em Dis} and {\em Gen} indicate a discriminative approach and a generative approach, respectively.
    {\em All}, {\em Head} and {\em Prompt} indicate finetuning all parameters of the LM, head tuning and prompt tuning, respectively.
    {\em Direct++ Demon} and {\em Channel Demon} indicate demonstration-based methods (the best out of concat-based and ensemble-based is taken).
    Models are run 4 times for $K=\mathrm{full}$ (4 verbalizers) and 20 times for others (4 verbalizers and 5 data seeds).
    \ifshorten{}\else{Channel models are more competitive with smaller $K$; less competitive with larger $K$.}\fi
}
\label{fig:k_ablation}
\end{figure*}

%\vspace{-.12cm}
%\paragraph{Demonstration (Section~\ref{subsec:main-no-finetune-result}) vs. Tuning}
%\citet{logan2021cutting} claims that prompt tuning does not outperform the demonstration method, which we find is true in direct models.
%When the channel models are used, prompt tuning outperforms the demonstration method by 3\% on average.
%We note that gains are inconsistent across datasets, likely because $K$ is small, but the later ablation shows that the performance of finetuning methods rapidly increases as $K$ increases while the demonstration method is inherently hard to scale.

\subsection{Ablations}\label{subsec:ablation-result}

For the ablations, we report experiments on SST-2, MR, TREC and AGNews, using one train seed (instead of four), and four verbalizers and five data seeds (as in main experiments).

%\vspace{-.12cm}
%\paragraph{Varying the size of LMs}
%We vary the size of LMs and report the average and the worst-case accuracy in Figure~\ref{fig:size_ablation}. The trends---no matter the best performance is achieved by channel prompt tuning or direct head tuning---are fairly consistent across varying size of LMs.

\vspace{-.12cm}
\paragraph{Varying the number of training examples}
We vary the number of training examples ($K$) and report the average accuracy in Figure~\ref{fig:k_ablation}.
All methods achieve higher accuracy as $K$ increases.
While we confirm strong performance of channel prompt tuning with $K \leq 16$, head tuning outperforms channel head tuning when $K=64$.
When $K=\mathrm{Full}$, both direct prompt tuning and head tuning outperform channel prompt tuning.
We think this is because (1) training signals amplified by channel models~\citep{lewis2018generative} are more significant when $K$ is small, and (2) channel models are more beneficial when labels on the training data are imbalanced (confirmed in the next ablation), which is more likely to happen with smaller $K$.

\ifshorten{}\else{
It is also worth noting that our experiment with $K=\mathrm{Full}$ confirms the finding from \citet{lester2021power} that direct prompt tuning matches the performance of all finetuning---finetuning all parameters of the LM---while being much more parameter-efficient.
This only holds with $K=\mathrm{Full}$; in a few-shot setup, all finetuning significantly outperforms other methods. This contradicts traditional analysis that having less trainable parameters is better when the training data is scarce~\citep{ng2002discriminative}. It is likely because such analysis did not take into account language model pretraining, which gives supervision to the model yet is not the training data for an end task.}\fi

\begin{figure}[t]
\resizebox{\columnwidth}{!}{\includegraphics[trim=0em 0 0em 0,clip]{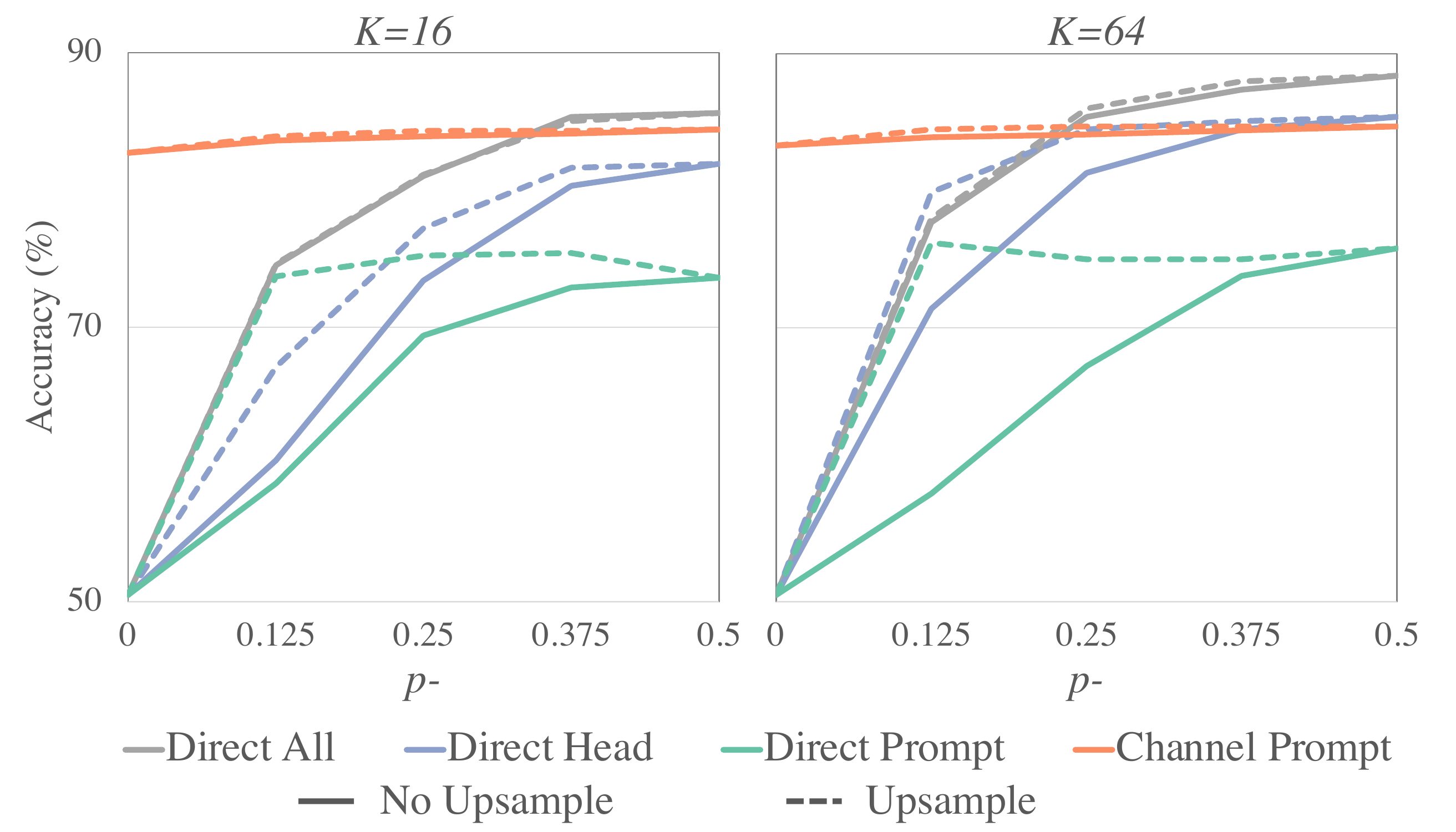}}
\caption{
    \textbf{Impact of imbalance in labels.} The average accuracy on SST-2 and MR of different methods with varying ratios of negative labels on the training data (denoted as $p^-$), when $K=16$ (left) or $64$ (right). As $p^-$ increases, the data is more balanced. \ifshorten{}\else{Channel models are more robust to imbalanced training data.}\fi
}
\label{fig:balance_ablation}
\end{figure}

\vspace{-.12cm}
\paragraph{Impact of imbalance in labels}
On binary datasets (SST-2 and MR), we vary the label imbalance in the training data with $K=\{16,64\}$. 
\ifshorten{
Specifically, let $\mathcal{C}=\{c^+, c^-\}$ and $p^-$ the ratio of $c^-$ in the training data,
we vary $p^-$ from $0$ to $0.5$. $p^-=0.5$ means the labels are perfectly balanced, and $p^-=0$ means labels in the training data only include $c^+$.
We additionally compare with {\em upsampling} baselines where we upsample training examples with infrequent labels so that the model has seen an equal number of examples per label during training.
}\else{
Specifically, let $\mathcal{C}=\{c^+, c^-\}$ and $p^-=|\{(x, c) \in \mathcal{D} | c=c^-\}|/|\mathcal{D}|$, i.e., the ratio of $c^-$ in the training data.
We vary $p^-$ to be $\{0, 0.125, 0.250, 0.375, 0.5\}$. $p^-=0.5$ means the labels are perfectly balanced, and $p^-=0$ means that labels in the training data only include $c^+$.
We additionally compare with {\em upsampling} baselines where we upsample training examples with infrequent labels so that the model has seen an equal number of examples per label during training.
}\fi

\providecommand{\blue}[1]{{\protect\color{blue!80!orange}{#1}}}

\begin{table*}[t]
    \centering \footnotesize
    \begin{tabular}{
            l @{\hspace{2em}} cc @{\hspace{2em}} ccccc
        }
        \toprule
            \multirow{2}{*}{Data} & \multicolumn{2}{c}{Zero-shot} & \multicolumn{5}{c}{Finetuning} \\
            \cmidrule(lr){2-3} \cmidrule(lr){4-8}
            %& Dr++ & Ch & Dr All & Dr Head & Dr Trans & Dr Prompt & Ch Prompt  \\
            & Direct++
            & Channel 
            & \makecell[c]{Direct \\  All}
            & \makecell[c]{Direct \\  Head}
            & \makecell[c]{Direct \\  Trans}
            & \makecell[c]{Direct \\  Prompt}
            & \makecell[c]{Channel \\  Prompt}
            \\
        \midrule    
            SST-2   & 80.3/76.9 & 77.1/74.8 &
            50.2/49.1 & 50.2/49.1 & 50.2/49.1 & 50.2/49.1 & \textbf{85.5}/\textbf{82.5} \\
            SST-5   & 33.3/28.8 & 29.2/27.7 & 
            \textbf{40.1}/\textbf{34.8} & 34.3/28.0 & 32.6/24.5 & 30.0/18.1 & {37.5}/{32.6} \\
            MR      & 77.4/73.2 & 74.3/69.3 & 
            50.0/50.0 & 50.0/50.0 & 50.0/50.0 & 50.0/50.0 & \textbf{80.9}/\textbf{74.8} \\
            CR      & 77.9/69.7 & 65.8/60.2 & 
            50.0/50.0 & 50.0/50.0 & 50.0/50.0 & 50.0/50.0 & \textbf{80.9}/\textbf{74.8} \\
            TREC    & 27.7/12.6 & 30.5/19.4 &
            \textbf{50.8}/31.0 & 44.8/29.6 & 44.6/\textbf{32.8} & 33.9/17.4 & 34.3/26.0 \\
            Subj    & 52.0/48.8 & 57.8/51.5 & 
            50.0/50.0 & 50.0/50.0 & 50.0/50.0 & 50.0/50.0 & \textbf{66.6}/\textbf{57.6} \\
        \bottomrule
    \end{tabular}
    \caption{
        Model performance when there is at least one label at test time that was unseen during training.
        All models are run 20 times (4 verbalizers and 5 data seeds).
        {\em All}, {\em Head}, {\em Trans} and {\em Prompt} indicate finetuning all parameters of the LM, head tuning, transformation tuning and prompt tuning, respectively.
        %{\em Dis} and {\em Gen} indicate discriminative approach and generative, respectively.
        We report the average and the worst-case accuracy, separated by a slash.
    }\label{tab:unseen-within-task}
\end{table*}

\begin{figure*}[t]
\resizebox{2.1\columnwidth}{!}{\includegraphics{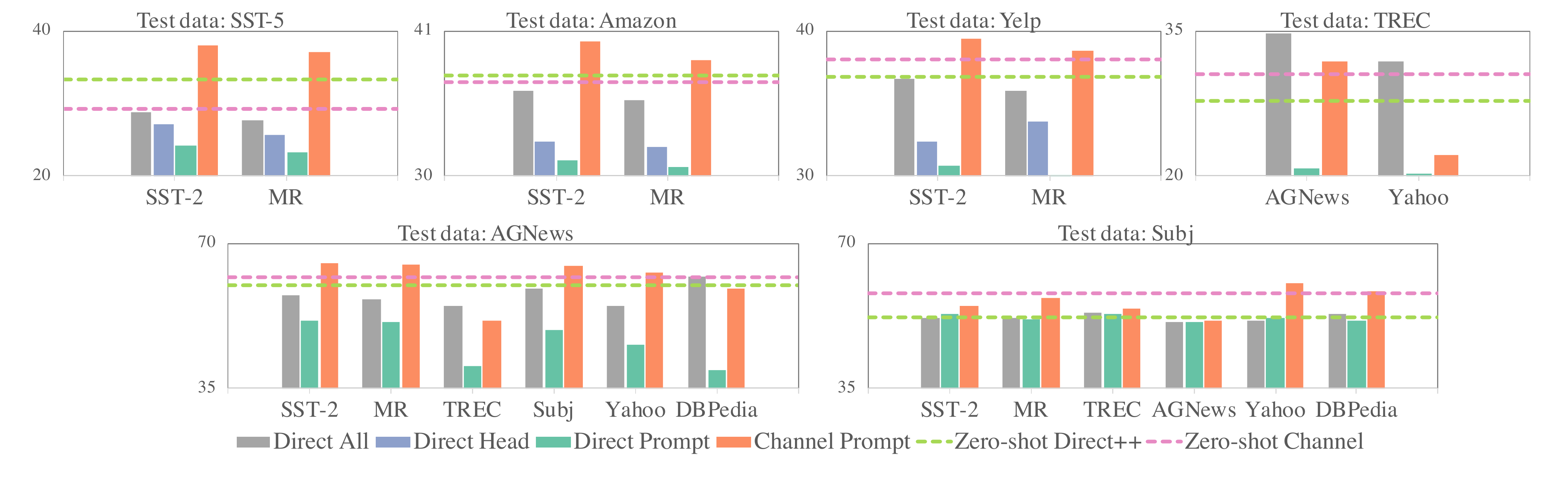}}
\caption{
    Model performance when transferred to unseen data, where x-axis indicates training data.
    {\em Direct Head} is not applicable when label space is not shared (when test datasets are TREC, AGNews and Subj).
    \ifshorten{}\else{Channel models have better generalization capacity than direct models.}\fi
}
\label{fig:unseen-across-task}
\end{figure*}

Results are reported in Figure~\ref{fig:balance_ablation}.
All direct models are sensitive to the imbalance in training data, even though they benefit from upsampling when $p^-$ is small.
Channel prompt tuning is insensitive to the imbalance, and significantly outperforms direct models when $p^-$ is small; it even outperforms all finetuning when $p^-<0.25$.
When $p^-$ is near to 0.5, direct head tuning matches or outperforms channel prompt tuning.

\ifshorten{}\else{It is also worth noting that direct prompt tuning with upsampling matches or outperforms all finetuning and head tuning when $p^-$ is small.
%Furthermore, direct prompt tuning with upsampling matches or outperforms all finetuning and head tuning when $p^-$ is small. However, as we saw in Section~\ref{subsec:main-finetune-result}, the performance of direct models are sometimes comparable to or even worse than zero-shot performance, but when the training data is balanced, they are always better than zero-shot performance.
}\fi

\subsection{Generalization to unseen labels}\label{subsec:generalization-result}

We experiment with a challenging scenario where the model must generalize to unseen labels. While it may be seen as an extreme scenario, this is often a practical setting, e.g., the problem is defined with a set of labels but later an addition of the new label may be needed.

First, we sample $K$ training examples as in main experiments but excluding one random label, so that at least one label at test time was unseen during training. Table~\ref{tab:unseen-within-task} reports the results. All direct models are unable to predict the label that is unseen at training time.
However, channel prompt tuning can predict unseen labels and achieves considerably better performance than zero-shot.
It outperforms all finetuning on 2-way classification datasets, and outperforms head tuning on five datasets except for TREC on which head tuning achieves very strong performance on seen labels.

Next, we run zero-shot transfer learning, where the model is trained on one dataset and is tested on another dataset.
Here, head tuning is not applicable when the labels are not shared between two datasets.
Figure~\ref{fig:unseen-across-task} shows the results. Channel prompt tuning outperforms all direct models including all finetuning on all datasets except for TREC.
It is particularly competitive when the tasks are inherently similar, e.g., transfer between 2-way sentiment analysis and 5-way sentiment analysis in the first three figures.
In fact, in such cases, performance is close to the models trained on in-domain data. %: 36.3, 43.4 and 43.9 vs. 38.0, 40.2 and 39.5 on SST-5, Amazon and Yelp, respectively.
When tasks are inherently different, e.g., the rest of the figures in Figure~\ref{fig:unseen-across-task}, gains over zero-shot performance are relatively small; we think more work should be done to make cross-task transfer better and to discover when it is possible.

\ifshorten{
\section{Conclusion}\label{sec:discuss}In this work, we introduced a noisy channel approach for few-shot text classification through LM prompting, where we either provide demonstrations to the LM or tune the prompt embeddings in the continuous space. 
Our experiments on eleven datasets show that channel models significantly outperform their direct counterparts, mainly because of their stability, i.e., lower variance and better worst-case accuracy.
We also found that direct head tuning is more competitive than previously thought, and different methods are preferred given different conditions. Specifically, channel prompt tuning is preferred in the following scenarios.

\vspace{-.1cm}
\paragraph{\bolden{$K$} is small}
Channel prompt tuning is more competitive when there are fewer training examples.
We hypothesize two reasons:
(1) Channel models are more stable (i.e., achieve low variance and high worst-case accuracy), unlike direct models that are highly unstable with small $k$~\citep{zhao2021calibrate,perez2021true,lu2021fantastically}.
(2) Channel models provide more signals by requiring the model to explain the input word-by-word (as claimed in \citet{lewis2018generative}) which is beneficial in the low data regime.
%This is in line with previous work: analysis in \citet{ng2002discriminative} shows that channel models are better when there are fewer training examples; \citet{lewis2018generative} claims that channel models provide more signals by requiring the model to explain the input word-by-word, which would be beneficial in the low data regime.

\vspace{-.1cm}
\paragraph{Data is imbalanced or \bolden{$|\mathcal{C}|$} is large}
When the training data is even slightly imbalanced, no direct models are competitive. We think this is because the LM head relies too much on unconditional distributions of labels. Channel prompt tuning is less sensitive because labels are only a conditioning variable.
Label imbalance in the training data is a real-world problem, especially when $k$ is small and $|\mathcal{C}|$ is large. We thus suggest this is an important area for future work.

\vspace{-.1cm}
\paragraph{Generalization to unseen labels is required}
All direct models are unable to predict labels that are unseen during training, indicating that they overfit in the label space. In contrast, channel models can predict unseen labels, likely because the label space is indirectly modeled.
This is in line with prior work that shows channel models are more competitive under a distribution shift~\citep{yogatama2017generative,lewis2018generative}.

\vspace{-.1cm}
\paragraph{Task is closer to language modeling}
If the task is too different from language modeling even with carefully chosen verbalizers (e.g., TREC and Subj), head tuning outperforms prompt tuning. This is likely because it benefits from directly updating the parameters of the LM. This may mean that causal LMs are not suitable for all tasks, or we need more sophisticated methods to apply causal LMs for such tasks without updating the LM parameters.

\vspace{.1cm}
\paragraph{Limitations and future work}
While we show that channel models are competitive in few-shot text classification, there are limitations that provide avenues for future work.
First, it is not as easy to use channel models for non classification tasks where modeling prior distributions is non-trivial. We think future work can obtain the prior with a separate model and incorporate it to the conditional LM as done by \citet{lewis2018generative}, potentially with beam search decoding as in \citet{yu2016neural,yee2019simple}.

Second, while this paper focuses on causal LMs, it is an open question how to use a channel model with masked LMs. Although we think channel models are not inherently restricted to causal LMs, the specific way in which existing masked LMs are pretrained makes it hard to use channel models without updating the LM parameters, e.g.,
masked LMs are not trained to generate long sentences.
One recent approach uses a label-conditioning objective \citep{tam2021improving} as a clever way to introduce a channel-like model with existing masked LMs.
Extending and further integrating these different approaches would be important for using channel models in a wider range of scenarios.
}
\else{
\section{Discussion \& Conclusion}\label{sec:discuss}
}\fi

\section*{Acknowledgements}
We thank Ari Holtzman, Eric Wallace, Gabriel Ilharco, Jungsoo Park, Myle Ott, Peter West and Ves Stoyanov for their helpful comments and discussion.
This research was supported by NSF IIS-2044660, ONR N00014-18-1-2826, 
an Allen Distinguished Investigator Award, and a Sloan Fellowship. 

\bibliography{acl}
\bibliographystyle{acl_natbib}

\clearpage
\appendix
\section{Samples \& Verbalizers}\label{app:verb}

Table~\ref{tab:data-samples} shows samples from each dataset.
Table~\ref{tab:verb} shows a list of verbalizers (four for each dataset), mainly taken from \citet{gao2021making} and label words included in the original data.

\newcommand{\mask}{\texttt{MASK}}

\begin{table*}[!b]
    \centering \footnotesize
    \begin{tabular}{l l l }
        \toprule
           Dataset & Verbalizers \\
        \midrule
            SST-2, MR &
            A \mask\ one.; It was \mask.; All in all \mask.; A \mask\ piece. (\mask=\{great, terrible\})\\
        \midrule
            SST-5, Amaon, Yelp &
            (Same as above.) (\mask=\{great,good,okay,bad terrible\})
            \\
        \midrule
            TREC & \makecell[l]{\mask: ; Q: \mask: ; Why \mask? ; Answer: \mask \\
            (\mask=\{Description, Entity, Expression, Human, Location, Number\}) } \\
        \midrule
            AGNews & \makecell[l]{Topic: \mask.; Subject: \mask.; This is about \mask.; It is about \mask. \\
            (\mask=\{World, Sports, Business, Technology\})} \\
        \midrule
            Yahoo & \makecell[l]{ (Same as above) (\mask=\{Society \& Culture, Science \& Mathematics, Health, Education \& \\ Reference, Computers \& Internet, Sports, Business \& Finance, Entertainment \& Music, \\ Family \& Relationships, Politics \& Government\})} \\
        \midrule
            DBPedia &\makecell[l]{ (Same as above) (\mask=\{Company, Educational Institution, Artist, Athlete, Office Holder, Mean \\ of Transportation, Building, Natural Place, Village, Animal, Plant, Album, Film, Written Work\})} \\
        \midrule
            Subj & This is \mask.; It's all \mask.' It's \mask.; Is it \mask? (\mask=\{subjective, objective\}) \\
        \bottomrule
    \end{tabular}\vspace{-.3em}
    \caption{
        Four different verbalizers for each dataset used in the experiments, separated by `;'. Verbalizers are taken from \citet{gao2021making} and label words included in the original data.
    }\label{tab:verb}
\end{table*}

\begin{table}[t]
    \centering \footnotesize
    \setlength{\tabcolsep}{0.2em}
    \begin{tabular}{l rrrr}
        \toprule
           \multirow{2}{*}{Data} & 
            \multicolumn{3}{c}{Direct} & Channel \\
            \cmidrule(lr){2-4} \cmidrule(lr){5-5}
            & Head & Trans & Prompt & Prompt \\
        \midrule
            SST-2, SST-5   & 0.001 & 0.001 & 0.01 & 0.001\\
            %SST-5   & 0.001 & 0.001 & 0.01 & 0.001\\
            MR      & 0.001 & 0.001 & 0.01 & 0.1\\
            CR      & 0.001 & 0.001 & 0.01 & 0.001 \\
            Amazon  & 0.001 & 0.001 & 0.001 & 0.1 \\
            Yelp    & 0.001 & 0.001 & 0.001 & 0.01\\
            TREC    & 0.001 & 0.001 & 0.01 & 0.01 \\
            AGNews  & 0.001 & 0.001 & 0.01 & 0.1 \\
            Yahoo   & 0.001 & 0.001 & 0.01 & 0.001 \\
            DBPedia & 0.001 & 0.001 & 0.01 & 0.01 \\
            Subj    & 0.001 & 0.001 & 0.01 & 0.01 \\
        \bottomrule
    \end{tabular}\vspace{-.3em}
    \caption{Learning rates of the models in Table~\ref{tab:finetune}.
    }\label{tab:lr}
\end{table}

\begin{table}[t]
    \centering \footnotesize
    \setlength{\tabcolsep}{0.2em}
    \begin{tabular}{l l rrr}
        \toprule
           \multirow{2}{*}{Data} & 
           \multirow{2}{*}{Size} & 
            \multicolumn{2}{c}{Direct} & Channel \\
            \cmidrule(lr){3-4} \cmidrule(lr){5-5}
            & & Head & Prompt & Prompt \\
        \midrule
            SST-2   & S,M,XL & 0.001 & 0.01 & 0.001\\
            MR      & S,M,XL & 0.001 & 0.01 & 0.1\\
            TREC & S    & 0.01 & 0.01 & 0.1 \\
            TREC & M & 0.01 & 0.01 & 1.0\\
            TREC & XL & 0.001 & 0.01 & 0.1 \\ 
            AGNews & S & 0.001 & 0.01 & 0.1 \\
            AGNews & M & 0.001 & 0.01 & 0.01 \\
            AGNews & XL & 0.001 & 0.01 & 0.001 \\
        \bottomrule
    \end{tabular}\vspace{-.3em}
    \caption{Learning rates of the models in Figure~\ref{fig:size_ablation}.
    }\label{tab:lr-size}
\end{table}

\begin{table}[t]
    \centering \footnotesize
    \setlength{\tabcolsep}{0.2em}
    \begin{tabular}{l l rrr}
        \toprule
            \multirow{2}{*}{Data} & 
           \multirow{2}{*}{$k$} & 
            \multicolumn{2}{c}{Direct} & Channel \\
            \cmidrule(lr){3-4} \cmidrule(lr){5-5}
            & & Head & Prompt & Prompt \\
        \midrule
            SST-2   & 4 & 0.001 & 0.001 & 0.001 \\
            SST-2   & 64 & 0.001 & 0.01 & 0.001 \\
            SST-2   & $\mathrm{Full}$ & 0.001 & 0.01 & 0.1 \\
            MR      & 4 & 0.001 & 0.001 & 0.001 \\
            MR      & 64,$\mathrm{Full}$ & 0.001 & 0.01 & 0.1 \\
            TREC    & 4 & 0.001 & 0.001 & 0.001 \\
            TREC    & 64,$\mathrm{Full}$ & 0.001 & 0.01 & 0.1 \\
            AGNews  & 4 & 0.001 & 0.001 & 0.1 \\
            AGNews  & 64 & 0.001 & 0.01 & 0.01 \\
            AGNews  & $\mathrm{Full}$ & 0.001 & 0.01 & 0.1 \\
        \bottomrule
    \end{tabular}\vspace{-.3em}
    \caption{Learning rates of the models in Figure~\ref{fig:k_ablation}.
    }\label{tab:lr-k}
\end{table}

\begin{table*}[t]
    \centering \footnotesize %\setlength{\extrarowheight}{-.2em}
    \begin{tabular}{l}
        \toprule
            {\em Data: SST-2, SST-5 and MR (Movie Sentiment Analysis)} \\
            \tabitem A three-hour cinema master class. ($c=$terrible) \\
            \tabitem A pretensions -- and disposable story --- sink the movie. ($c=$great) \\
        \cmidrule(lr){1-1}
            {\em Data: CR} \\
            \tabitem It is slow, if you keep the original configuration and prigs (why'd u buy it then?!) it'll run smoothly, but still slower \\ ~~~~then
            most other coloured-screen nokias.  ($c=$terrible)\\
            \tabitem It takes excellent pics and is very easy to use, if you read the manual. ($c=$great) \\
        \cmidrule(lr){1-1}
            {\em Data: Amazon} \\
            \tabitem Don't waste your money if you already have 2003... There isn't one reason to get this update if you already have MS \\ ~~~~Money 2003 Deluxe and Business.  ($c=$terrible)\\
            \tabitem The game was in perfect condition! came before it said it should have by 2 days!! I love the game and I suggest it to \\ ~~~~a lot of my friends!! ($c=$great)
            \\
        \cmidrule(lr){1-1}
            {\em Data: Yelp} \\
            \tabitem I've eaten at the other location, and liked it. But I tried this place, and I have JUST NOW recovered physically enough \\ ~~~~from the worst food poisoning I've ever heard of to write this review. ($c=$terrible)\\
            \tabitem Great ambiance, awesome appetizers, fantastic pizza, flawless customer service. ($c=$great)\\
        \cmidrule(lr){1-1}
            {\em Data: TREC} \\
            \tabitem How do you get a broken cork out of a bottle? ($c=$Description) \\
            \tabitem Mississippi is nicknamed what? ($c=$Entity) \\
            \tabitem What is BPH? ($c=$Expression) \\
            \tabitem Who won the Novel Peace Prize in 1991? ($c=$Human) \\
            \tabitem What stadium do the Miami Dolphins play their home games in? ($c=$Location) \\
            \tabitem How long did the Charles Manson murder trial last? ($c=$Number)
            \\
        \cmidrule(lr){1-1}
            {\em Data: AGNews} \\
            \tabitem Peru Rebel Leader Offers to Surrender Reuters - The leader of an armed group which took over a police station in a \\ ~~~~southern Peruvian town three days ago and demanded the president's resignation ... ($c=$World) \\
            \tabitem Walk in park for Yankees Drained by a difficult week, the New York Yankees needed an uplifting victory. ($c=$Sports)\\
            \tabitem Schwab plans new, smaller branches SAN FRANCISCO -- Charles Schwab \& Co. is opening new offices that are \\ ~~~~smaller than its current branches ... ($c=$Business) \\
            \tabitem NASA Mountain View claims world's fastest computer. ($c=$Technology)
            \\
        \cmidrule(lr){1-1}
            {\em Data: Yahoo} \\
            \tabitem What's one change you could make to your lifestyle that would give you more peace? ... ($c=$Society \& Culture)\\
            \tabitem If the average for a test was 74\% and the standard deviation was 13, are you within 1 SD if you scored a 62? \\ ~~~~($c=$Science \& Mathematics)\\
            \tabitem Can someone explain to me what IndexOf is in Visual Basic? ($c=$Computers \& Internet)
            \\
        \cmidrule(lr){1-1}
            {\em Data: DBPedia} \\
            \tabitem Coca-Cola Bottling Co. Consolidated headquartered in Charlotte North Carolina is the largest independent Coca- \\ ~~~~Cola bottler in the United States ... ($c=$Company)\\
            \tabitem Elk County Catholic High School is a private Roman Catholic high school in ... ($c=$Educational Institution)\\
            \tabitem Louis Wiltshire (born 23 April 1969) is a British sculptor. ... ($c=$Artist) \\
            \tabitem Russel Paul Kemmerer (botn November 1 1931 in Pittsburgh Pennsylvania) is an American retired professional \\ ~~~~baseball player. ($c=$Athlete) \\
            \tabitem Dialectica aemula is a moth of the Gracillariidae family. ... ($c=$Animal) \\
            \tabitem Ephedra viridis known by the common names green Mormon tea green ephedra is a species of Ephedra. ($c=$Plant)\\
        \cmidrule(lr){1-1}
            {\em Data: Subj} \\
            \tabitem As i settled into my world war ii memory, i found myself strangely moved by even the corniest and most hackneyed \\ ~~~~contrivances. ($c=$subjective) \\
            \tabitem This is a story about the warm relationship between a little girl and her father despite the difficult conditions they \\ ~~~~have to live in. ($c=$objective)\\
        \bottomrule
    \end{tabular}
    \caption{
        Samples from each dataset. $c$ indicates the label.
    }\label{tab:data-samples}
\end{table*}

\section{Implementation Details}\label{app:hyperparams}

We use PyTorch~\citep{paszke2019pytorch} and Huggingface Transformers~\citep{wolf-etal-2020-transformers}. For MR, we use the sentence polarity dataset version 1.0.
We use the batch size of 32 and the sequence length of 128 for datasets with short input text (SST-2, SST-5, MR, TREC) and the batch size of 16 and the sequence length of 256 for datasets with long input text (AGNews, Amazon, Yelp, DBPedia, Yahoo, Subj). When the concat-based demonstration method is used, the sequence length is multiplied by the number of training examples, yet is bounded by 1024 which is a strict limit of GPT-2.

For all finetuning experiments, we train the model for 100 global steps. We use the loss divided by the number of all tokens in the batch.
We use Adam optimizer~\citep{kingma2015adam} with no weight decay and no warmup steps.
For head tuning, transformation tuning and prompt tuning, we use the learning rate $\{0.1, 0.01, 0.001\}$ and choose the one that gives the lowest training loss on average in order to eliminate the need of the validation data. The chosen learning rate values are reported in Table~\ref{tab:lr}.
For all finetuning, we use the learning rate of $10^{-5}$.
%For a full-shot setup which we experiment with for checking if we have successfully reproduced the original prompt-tuning method, we use the full validation set.
For prompt tuning, we use $n=20$ prompt tokens which embeddings are initialized from a random subset of the top $5000$ vocabularies, following the original paper~\citep{lester2021power}.

\begin{table*}[t]
    \centering \footnotesize
    \setlength{\tabcolsep}{0.2em}
    \begin{tabular}{
            l
            @{\hspace{3.0em}}llr
            @{\hspace{3.0em}}llr
            @{\hspace{3.0em}}llr
        }
        \toprule
            \multirow{2}{*}{Data} &
            \multicolumn{3}{c}{Direct} &
            \multicolumn{3}{c}{Direct++} & \multicolumn{3}{c}{Channel} \\
            \cmidrule(lr){2-4} \cmidrule(lr){5-7} \cmidrule(lr){8-10}
            & Avg$_\text{(Std)}$ & Best & Worst
            & Avg$_\text{(Std)}$ & Best & Worst
            & Avg$_\text{(Std)}$ & Best & Worst
            \\
        \midrule    
    SST-2 &
    58.9$_{(9.4)}$ & 77.4 & 50.6 & 66.8$_{(8.2)}$ & 81.0 & 51.7 & \textbf{85.0}$_{(1.1)}$ & 86.9 & \textbf{83.1}
    \\
    SST-5 &
    27.6$_{(5.2)}$ & 40.9 & 23.0 & 23.7$_{(4.5)}$ & 31.4 & 14.4 & \textbf{36.2}$_{(2.1)}$ & 39.6 & \textbf{32.7}
    \\
    MR & 
    56.4$_{(8.5)}$ & 78.2 & 50.0 & 60.2$_{(8.6)}$ & 79.0 & 50.5 & \textbf{80.5}$_{(1.8)}$ & 83.2 & \textbf{76.8}
    \\
    CR &
    54.7$_{(7.9)}$ & 78.8 & 50.0 & 66.8$_{(9.8)}$ & 84.0 & 50.0 & \textbf{80.8}$_{(3.3)}$ & 86.2 & \textbf{74.8} 
    \\
    Amazon &
    33.0$_{(6.5)}$ & 43.6 & 21.4 & \textbf{40.8}$_{(2.5)}$ & 46.4 & 35.7 & 39.4$_{(2.5)}$ & 42.6 & 34.3
    \\
    Yelp &
    32.6$_{(5.1)}$ & 41.6 & 23.3 & 38.5$_{(3.6)}$ & 44.0 & 31.6 & 39.8$_{(2.1)}$ & 43.8 & 36.5 
    \\
    AGNews &
    34.0$_{(10.9)}$ & 62.3 & 25.0 & 51.2$_{(10.2)}$ & 68.0 & 34.4 & 68.5$_{(4.5)}$ & 76.1 & 60.6
    \\
    TREC & 
    27.2$_{(9.2)}$ & 42.0 & 9.4 & 31.6$_{(18.9)}$ & \textbf{78.4} & 13.0 & \textbf{42.0}$_{(7.1)}$ & 54.4 & \textbf{26.8}
    \\
    Yahoo &
    13.0$_{(2.6)}$ & 18.7 & 10.0 & 29.6$_{(6.2)}$ & 40.7 & 19.4 & 56.2$_{(1.2)}$ & 57.7 & 52.3
    \\
    DBPedia & 
    32.5$_{(17.0)}$ & 68.2 & 7.1 & 71.1$_{(8.0)}$ & \textbf{82.4} & 55.2 & 58.5$_{(12.5)}$ & 74.3 & 40.0
    \\
    Subj &
    53.7$_{(6.0)}$ & 71.8 & 49.9 & 56.9$_{(8.2)}$ & \textbf{75.9} & 50.0 & \textbf{60.5}$_{(6.5)}$ & 68.0 & 40.8
    \\
    \midrule
    Avg. & 38.5 & 56.7 & 29.1 & 48.8 & 64.7 & 36.9 & \textbf{58.9} & 64.8 & \textbf{50.8}
    \\
        \bottomrule
    \end{tabular}
    \caption{
        Full results from demonstration methods when a \textbf{concat-based method} is used; analogous to Table~\ref{tab:main}.
        {\em Avg}, {\em Std}, {\em Best} and {\em Worst} indicate the average accuracy, standard deviation, the best-case accuracy and the worst-case accuracy, respectively.
        Bold: Best when combined with Table~\ref{tab:no-finetune-ensemble-full-result}.
    }\label{tab:no-finetune-concat-full-result}
\end{table*}

\begin{table*}[t]
    \centering \footnotesize
    \setlength{\tabcolsep}{0.2em}
    \begin{tabular}{
            l
            @{\hspace{3.0em}}llr
            @{\hspace{3.0em}}llr
            @{\hspace{3.0em}}llr
        }
        \toprule
            \multirow{2}{*}{Data} &
            \multicolumn{3}{c}{Direct} &
            \multicolumn{3}{c}{Direct++} & \multicolumn{3}{c}{Channel} \\
            \cmidrule(lr){2-4} \cmidrule(lr){5-7} \cmidrule(lr){8-10}
            & Avg$_\text{(Std)}$ & Best & Worst
            & Avg$_\text{(Std)}$ & Best & Worst
            & Avg$_\text{(Std)}$ & Best & Worst
            \\
        \midrule    
    SST-2 &
    57.5$_{(9.6)}$ & 84.2 & 50.9 & 79.7$_{(5.8)}$ & \textbf{88.3} & 68.0 & 77.5$_{(7.9)}$ & 85.9 & 59.5
    \\
    SST-5 &
    25.6$_{(2.7)}$ & 34.6 & 23.2 & 33.8$_{(5.8)}$ & \textbf{42.4} & 23.3 & 33.6$_{(2.2)}$ & 38.0 & 30.2
    \\
    MR & 
    58.8$_{(9.9)}$ & 82.9 & 50.0 & 76.8$_{(6.4)}$ & \textbf{85.7} & 60.1 & 76.1$_{(6.6)}$ & 82.0 & 60.0
    \\
    CR &
    51.0$_{(2.2)}$ & 59.0 & 50.0 & 72.8$_{(12.0)}$ & \textbf{87.4} & 54.6 & 79.7$_{(4.2)}$ & 84.0 & 69.3
    \\
    Amazon &
    31.7$_{(6.1)}$ & 44.5 & 23.1 & 39.8$_{(4.6)}$ & \textbf{47.8} & 32.0 & 40.4$_{(2.1)}$ & 44.3 & \textbf{36.2}
    \\
    Yelp &
    31.4$_{(6.3)}$ & 41.4 & 23.6 & 39.2$_{(6.1)}$ & \textbf{47.3} & 29.6 & \textbf{41.5}$_{(1.3)}$ & 43.5 & \textbf{38.5}
    \\
    AGNews &
    51.9$_{(9.8)}$ & 69.7 & 34.2 & 73.1$_{(6.2)}$ & \textbf{81.8} & 58.6 & \textbf{74.3}$_{(2.7)}$ & 78.5 & \textbf{69.3}
    \\
    TREC & 
    32.1$_{(10.4)}$ & 54.4 & 13.0 & 22.9$_{(10.1)}$ & 44.4 & 9.8 & 31.5$_{(5.0)}$ & 43.2 & 23.8
    \\
    Yahoo &
    16.6$_{(4.2)}$ & 24.6 & 10.7 & 50.6$_{(2.1)}$ & 54.1 & 46.5 & \textbf{58.6}$_{(0.7)}$ & \textbf{59.7} & \textbf{57.4}
    \\
    DBPedia & 
    46.8$_{(15.2)}$ & 63.0 & 17.1 & \textbf{72.6}$_{(7.0)}$ & 81.9 & 55.7 & 64.8$_{(3.5)}$ & 70.0 & \textbf{57.0}
    \\
    Subj &
    51.6$_{(3.4)}$ & 62.3 & 49.6 & 52.2$_{(5.4)}$ & 61.8 & 41.8 & 52.4$_{(3.0)}$ & 57.7 & 46.9
    \\
    \midrule
    Avg. & 41.4 & 56.4 & 31.4 & 55.8 & \textbf{65.7} & 43.6 & 57.3 & 62.4 & 49.8 \\
    \bottomrule
    \end{tabular}
    \caption{
        Full results from demonstration methods when a \textbf{ensemble-based method} is used; analogous to Table~\ref{tab:main}.
        {\em Avg}, {\em Std}, {\em Best} and {\em Worst} indicate the average accuracy, standard deviation, the best-case accuracy and the worst-case accuracy, respectively.
        Bold: Best when combined with Table~\ref{tab:no-finetune-concat-full-result}.
    }\label{tab:no-finetune-ensemble-full-result}
\end{table*}

\begin{table*}[t]
    \centering \footnotesize
    \setlength{\tabcolsep}{0.2em}
    \begin{tabular}{
            l
            @{\hspace{3.0em}}llr
            @{\hspace{2.6em}}llr
            @{\hspace{2.6em}}llr
            @{\hspace{2.6em}}llr
        }
        \toprule
            \multirow{2}{*}{Data} &
            \multicolumn{3}{c}{Direct Head} &
            \multicolumn{3}{c}{Direct Trans} & \multicolumn{3}{c}{Direct Prompt} & \multicolumn{3}{c}{Channel Prompt} \\
            \cmidrule(lr){2-4} \cmidrule(lr){5-7} \cmidrule(lr){8-10} \cmidrule(lr){11-13}
            & Avg$_\text{(Std)}$ & Best & Worst
            & Avg$_\text{(Std)}$ & Best & Worst
            & Avg$_\text{(Std)}$ & Best & Worst
            & Avg$_\text{(Std)}$ & Best & Worst
            \\
        \midrule    
            SST-2 &
            80.2$_{(5.1)}$ & 88.4 & 68.6 & 77.3$_{(5.6)}$ & 87.7 & 57.5 & 72.6$_{(10.0)}$ & \textbf{89.3} & 50.9 & \textbf{85.8}$_{(1.5)}$ & 88.3 & \textbf{81.3} \\
            SST-5 &
            34.9$_{(2.8)}$&40.1&30.0 & 33.0$_{(2.7)}$&40.0&25.5 & 30.9$_{(5.8)}$&\textbf{42.6}&19.1 & \textbf{36.3}$_{(3.0)}$&41.6&\textbf{27.9} \\
            MR &
            73.7$_{(7.7)}$&83.9&\textbf{56.4} & 71.3$_{(8.1)}$&83.2&51.6 & 67.4$_{(9.9)}$&\textbf{85.1}&50.1 & \textbf{81.7}$_{(1.4)}$&84.2&78.0 \\
            CR &
            67.6$_{(10.5)}$&84.0&50.0 & 63.9$_{(9.6)}$&84.5&50.0 & 65.7$_{(13.2)}$&\textbf{87.4}&50.0 & \textbf{79.6}$_{(1.4)}$&82.7&\textbf{76.4} \\
            Amazon &
            34.5$_{(3.5)}$&41.4&28.8 & 32.1$_{(4.6)}$&40.2&18.2 & 31.2$_{(5.7)}$&43.6&20.0 & \textbf{43.4}$_{(2.3)}$&\textbf{49.2}&\textbf{39.2} \\
            Yelp &
            40.6$_{(4.0)}$&46.9&32.8 & 38.9$_{(3.3)}$&46.3&31.5 & 31.9$_{(7.7)}$&45.0&20.6 & \textbf{43.9}$_{(2.2)}$&\textbf{47.2}&\textbf{37.2} \\
            TREC &
            \textbf{54.1}$_{(7.1)}$&\textbf{71.2}&\textbf{42.4} & 48.0$_{(7.4)}$&66.6&31.0 & 35.9$_{(11.8)}$&65.8&13.0 & 37.1$_{(7.3)}$&55.8&20.8 \\
            AGNews &
            \textbf{74.1}$_{(6.6)}$&\textbf{84.5}&61.2 & 66.9$_{(8.0)}$&83.5&47.0 & 61.9$_{(15.9)}$&83.5&25.2 & 73.4$_{(3.1)}$&77.9&\textbf{63.9} \\
            Yahoo &
            39.1$_{(3.2)}$&44.9&31.4 & 33.8$_{(4.5)}$&43.8&23.0 & 27.4$_{(5.6)}$&39.0&15.7 & \textbf{54.0}$_{(2.0)}$&\textbf{57.6}&\textbf{46.7} \\ 
            DBPedia &
            49.3$_{(7.7)}$&64.2&37.5 & 42.4$_{(6.8)}$&56.9&28.6 & 41.8$_{(13.3)}$&75.3&9.9 & \textbf{67.7}$_{(5.7)}$&\textbf{78.3}&\textbf{52.9} \\
            Subj &
            \textbf{86.3}$_{(3.0)}$&\textbf{90.9}&\textbf{79.1} & 86.0$_{(4.0)}$&90.8&71.6 & 65.5$_{(7.7)}$&78.7&49.9 & 75.5$_{(5.0)}$&84.5&58.8 \\
        \midrule
            Avg. &
            57.7&67.3&47.1 & 54.0&65.8&39.6 & 48.4&66.9&29.5 & \textbf{61.7}&\textbf{67.9}&\textbf{53.0} \\
        \bottomrule
    \end{tabular}
    \caption{
        Full results from tuning methods; analogous to Table~\ref{tab:finetune}.
        {\em Head}, {\em Trans}, {\em Prompt} indicate head tuning, transformation tuning and prompt tuning, respectively.
        {\em Avg}, {\em Std}, {\em Best} and {\em Worst} indicate the average accuracy, standard deviation, the best-case accuracy and the worst-case accuracy, respectively.
        %`Avg.' in the last row indicates the macro-average across all datasets.
    }\label{tab:finetune-full-result}
\end{table*}

\clearpage
\section{Additional Results}\label{app:ablations}

\begin{figure*}[t]
\resizebox{2.1\columnwidth}{!}{\includegraphics{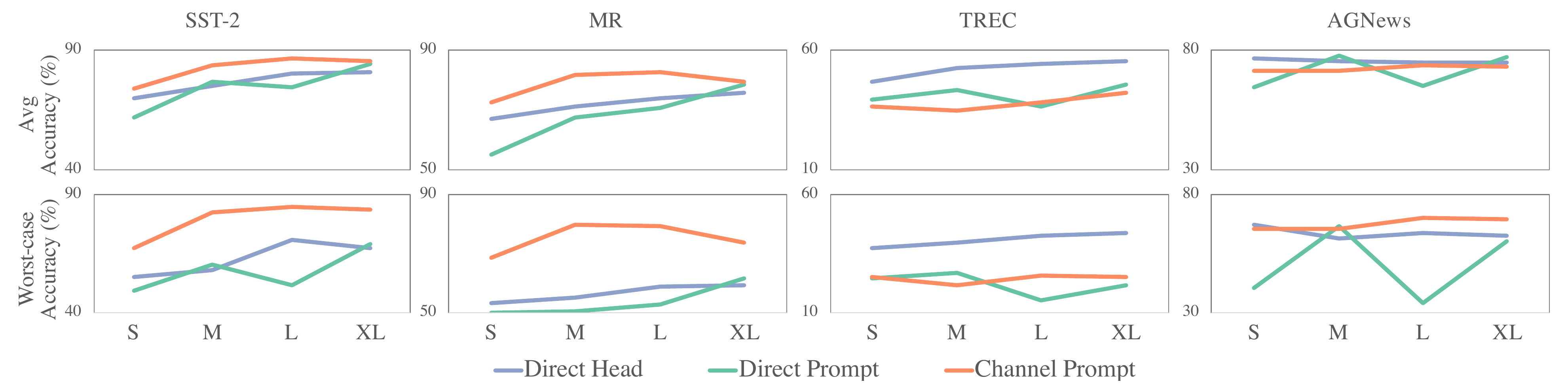}}
\caption{
    \textbf{Varying the size of LMs} from GPT-2 Small to GPT-2 X-Large.
    The average accuracy (top) and the worst-case accuracy (bottom) are reported.
    All models are run 20 times (4 verbalizers and 5 data seeds).
    %{\em Dis} and {\em Gen} indicate a discriminative approach and a generative approach, respectively.
    {\em Head} and {\em Prompt} indicate head tuning and prompt tuning, respectively.
    {Trends are consistent across different sizes of LM.}
}
\label{fig:size_ablation}
\end{figure*}

\paragraph{More metrics}
Table~\ref{tab:no-finetune-concat-full-result}, \ref{tab:no-finetune-ensemble-full-result} and \ref{tab:finetune-full-result} report the average accuracy, the variance, the best-case accuracy and the worst-case accuracy using the concat-based demonstration, the ensemble-based demonstration and the tuning methods, respectively.
Results consistently indicate that channel models achieve significantly lower variance and higher worst-case accuracy. The best-case accuracy is often achieved by direct models, but channel models outperform direct models on average.

\paragraph{Varying the size of LMs}
We vary the size of LMs and report the average and the worst-case accuracy in Figure~\ref{fig:size_ablation}.
The trends---no matter the best performance is achieved by channel prompt tuning or direct head tuning---are fairly consistent across varying size of LMs.

\end{document}